\documentclass[11pt]{article}

\usepackage[preprint]{acl}

\usepackage{times}
\usepackage{latexsym}

\usepackage[T1]{fontenc}
\usepackage[utf8]{inputenc}

\usepackage{microtype}

\usepackage{inconsolata}

\usepackage{graphicx}
\usepackage{amsmath,amssymb,amsfonts}
\usepackage{algorithm}
\usepackage{algpseudocode}
\usepackage{booktabs}
\usepackage{multirow}
\usepackage{makecell}
\usepackage[table]{xcolor}
\usepackage{graphicx}
\usepackage{pifont}
\usepackage{tabularx}
\usepackage{array}
\usepackage{arydshln}
\usepackage{enumitem}
\usepackage{subcaption}
\usepackage{needspace}

\newcommand{\dashedmidrule}{%
  \noalign{\vskip\aboverulesep}
  \cdashline{1-12}
  \noalign{\vskip\belowrulesep}
}

\newcommand{\cmark}{\checkmark}
\newcommand{\xmark}{\ding{55}}

\usepackage{xcolor}
\usepackage{listings}
\usepackage[most]{tcolorbox}

\usepackage{cleveref}
\title{Detector-Evasive LLM Paraphrasing via Constrained Policy Optimization}
\author{
Mingyi Wang$^1$ \quad
Zhuoer Shen$^2$ \quad
Yuheng Bu$^2$ \quad
Shaofeng Zou$^1$\thanks{Corresponding author.} \\
$^1$School of ECEE, Arizona State University \\
$^2$Department of Computer Science, University of California, Santa Barbara \\
\texttt{mwang287@asu.edu, zhuoershen@ucsb.edu} \\
\texttt{buyuheng@ucsb.edu, zou@asu.edu}
}

\begin{document}
\maketitle
\begin{abstract}

AI-text detectors are vulnerable to paraphrasing and detector-guided paraphrasing attacks, but existing detector-evasion methods often lack precise control over semantic preservation. In particular, optimizing directly for detector evasion can degrade fine-grained semantics, whereas scalarized reward designs provide only indirect, weight-sensitive control over the evasion–semantics trade-off. We address this limitation by formulating detector-evasive LLM paraphrasing as a Constrained Markov Decision Process, where detector evasion is the primary objective and semantic preservation is enforced as an explicit constraint. We propose Detector Evasion Policy Optimization (DEPO), a Lagrangian primal-dual reinforcement learning algorithm with a novel GRPO-style group-based policy update. DEPO adaptively balances semantic preservation and detector evasion during training, enabling the policy to improve attack success within a prescribed semantic-preservation region. Experiments on MAGE, M4, RAID, and peer-review datasets, evaluated against MAGE, RoBERTa, RADAR, Binoculars, and Fast-DetectGPT detectors, show that DEPO achieves strong detector evasion while precisely satisfying the semantic preservation constraint. DEPO also exhibits cross-domain, cross-detector, and prompt-level robustness. 
%Additional analysis shows that BERTScore is better suited than T5 sentence-level similarity for capturing structural and logic-sensitive fine-grained semantic changes.
\end{abstract}

\section{Introduction}

As large language models (LLMs) are increasingly adopted across diverse domains, identifying AI-generated text has become an important and timely challenge due to concerns about its misuse, e.g., academic dishonesty~\cite{perkins2023game} and spread of fake news~\cite{hanley2024machine}. This demand has led to the development of various AI-text detectors to distinguish machine-generated content from human-written text, e.g., \cite{wu2025survey,yang2024survey,kumarage2024survey,kehkashan2025ai}. 
Nevertheless, AI text detectors are inherently vulnerable, and their predictions can be severely compromised by various counter-detection strategies, e.g., token-level perturbations, paraphrasing-based rewriting, detector-guided adversarial paraphrasing, and homoglyph substitutions~\cite{liu2025robustness,huang2024robust,zhou2024humanizing,cheng2026adversarial,creo2025silverspeak}. Therefore,
studying attacks against these detectors is urgent, as it exposes the vulnerabilities of existing detection systems and provides stress tests that can inform the design of more robust detectors~\cite{sadasivan2025can}. 

% Among these attack methods, paraphrasing-based methods are particularly important because they directly correspond to the practical setting in which AI-generated text is revised to evade detection while preserving its semantics~\cite{krishna2023paraphrasing,sadasivan2025can,cheng2026adversarial}.% It is therefore of key importance to understand the limitations of existing detectors and provide insights for building more robust detection systems~\cite{sadasivan2025can}.

We categorize existing detector-evasion attacks based on how they exploit the detector as follows. The first class of methods does not use any feedback from the detector and operates at the word-level. Specifically, they reduce detector confidence through local character-level or token-level perturbations, such as adversarial substitutions or homoglyph-based modifications, and do not fully rewrite the text~\cite{huang2024robust,creo2025silverspeak}. 
The second class of methods uses LLMs to paraphrase or rewrite the generated content. Specifically, detector feedback is used to construct supervised fine-tuning datasets \cite{krishna2023paraphrasing}, as guided sampling score \cite{cheng2026adversarial,sadasivan2025can}.
%Other methods operate at the rewriting level, using paraphrasing or adversarial rewriting to make AI-generated text appear more human-written while preserving its original meaning~\cite{krishna2023paraphrasing,sadasivan2025can,cheng2026adversarial}.
The third class is reinforcement learning (RL)-based approaches, which directly fine-tune a paraphrasing LLM with detector feedback, where lower detector confidence or higher human-likeness scores are used as rewards to guide policy optimization~\cite{david2025authormist,ranganath2026stealthrl}. 
\begin{figure*}[!h]
    \centering
    \includegraphics[width=0.9\textwidth]{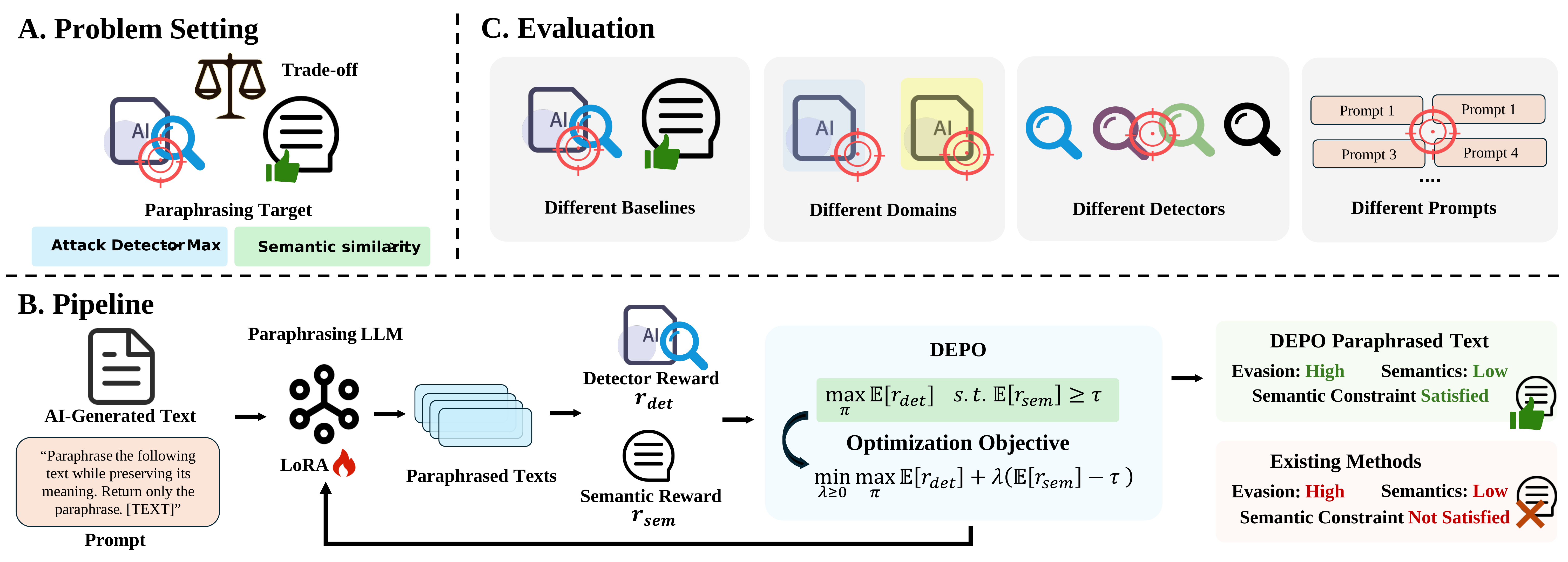}
    \caption{The DEPO framework for detector-evasive paraphrasing.(A) Problem setting: maximizing detector evasion $\mathbb{E}[r_{\mathrm{det}}]$ under the semantic constraint $\mathbb{E}[r_{\mathrm{sem}}]\geq\tau$.
    (B) Pipeline: DEPO uses detector and semantic rewards to optimize a paraphrasing LLM with constrained policy optimization.
    (C) Evaluation: DEPO also exhibits cross-domain, cross-detector, and prompt-level robustness.
    }
    \label{fig:main_pipeline}
\end{figure*}

Despite their success in attacking the detector, a central trade-off has not yet been adequately addressed: a successful attack should not merely reduce the detector score, but also preserve the fine-grained semantics of the original input. As we will show in our experiments, existing approaches often improve detector evasion by sacrificing semantic fidelity, and even when semantic preservation is incorporated, they lack an explicit mechanism for precisely controlling semantic degradation.

% \vspace{-0.1cm}
\subsection{Main Contributions}
% \vspace{-0.1cm}
To address these limitations, we propose a constrained RL framework for detector-evasive LLM paraphrasing. Given AI-generated content as input, the paraphrasing LLM is optimized to generate a rewritten output that evades AI-text detectors while preserving the fine-grained semantics of the original text.
We formulate this problem as a Constrained Markov Decision Process (CMDP), where detector evasion is optimized as the primary objective and semantic preserving requirements are imposed as explicit constraints~\cite{altman1999constrained,achiam2017constrained,tessler2019reward,li2024faster}. Under this formulation, the model improves attack success only within a feasible region defined by semantic preservation. 

Building on this formulation, we introduce DEPO, short for Detector Evasion Policy Optimization, a Lagrangian primal-dual policy optimization algorithm that combines a group-based policy gradient with an in-group baseline, following GRPO \cite{shao2024deepseekmath}, in the policy update to reduce computational costs and improve stability. During training, DEPO dynamically adjusts dual variables to balance detector reward maximization with constraint satisfaction, avoiding the ambiguity of optimizing a single composite reward.
We note that prior RL-based detector-evasive methods use a scalarized reward design, where detector evasion and output semantic preservation are linearly combined with a manually chosen weight~\cite{david2025authormist,ranganath2026stealthrl}, which, however, does not provide precise semantic preservation (as will be shown in our experiments). 

% In multi-objective optimization, linear scalarization is a standard approach for selecting Pareto-optimal trade-off solutions, particularly supported solutions under suitable conditions~\cite{miettinen1999nonlinear,marler2010weighted}. However, such a formulation does not provide direct control over task-specific quality thresholds: changing the weights can alter the resulting trade-off in a non-transparent and task-dependent manner, and a policy optimized for the scalar reward may still violate semantic preservation requirements. Thus, while scalarized rewards are simple and effective, they are not well suited for specifying and maintaining an explicit quality-preserving region.

We perform extensive experiments across diverse mainstream datasets, including MAGE~\cite{li2024mage}, M4~\cite{wang2024m4}, RAID~\cite{dugan2024raid}, and a peer-review dataset \cite{yu2025peerreview}, and evaluations against MAGE detector~\cite{li2024mage}, RoBERTa OpenAI~\cite{solaiman2019release}, RADAR~\cite{hu2023radar}, Binoculars~\cite{hans2024spotting}, and Fast-DetectGPT~\cite{bao2024fastdetectgpt} detectors, and demonstrate DEPO's strong detector evasion with precise semantic preservation. 
We further demonstrate the cross-domain, cross-detector, and prompt-level generalization and robustness, showing our LLM paraphrasing generalizes to other detectors and domains, and is robust to variation in paraphrasing instructions.
Importantly, we employ a more fine-grained BERTScore~\cite{zhang2020bertscore} metric (token-level similarity) to measure semantic preservation and show its advantage over the T5 similarity (between sentence-level embeddings)~\cite{ni2022sentencet5,ranganath2026stealthrl}. 
\subsection{Related Work}
% \vspace{-0.05cm}

\textbf{AI Text Detection Methods.}
Benchmark datasets for AI-text detection were developed, including open-domain question answering, news, stories, encyclopedic text, peer review, and online reviews~\citep{guo2023hc3,li2024mage,yu2025peerreview,gambetti2023combat}. Further, binary classification has become a representative detection paradigm: pretrained language encoders such as RoBERTa~\cite{solaiman2019release} are fine-tuned on human- and AI-written corpora~\citep{zellers2019defending,ippolito2020automatic,uchendu2021turingbench,li2024mage}. Recent works strengthened it through adversarial training, content-oriented modeling, and contrastive learning~\citep{hu2023radar,chen2025coconuts,guo2024detective}.

Another line of work develops zero-shot or training-free detection methods. 
Early studies exploit statistical signatures, including token-rank distributions and likelihood patterns
\citep{gehrmann2019gltr}. DetectGPT uses probability curvature \citep{mitchell2023detectgpt}, while Fast-DetectGPT improves its efficiency through conditional probability curvature \citep{bao2024fastdetectgpt}. Other methods, such as DNA-GPT, Ghostbuster, and Binoculars, exploit $n$-gram divergence, learned feature combinations, or cross-model perplexity discrepancies \citep{yang2024dnagpt, verma2024ghostbuster, hans2024spotting}. In addition, watermarking methods embed detectable statistical patterns during generation \citep{kirchenbauer2023watermark}. Despite their effectiveness, these detectors remain vulnerable when adversaries intentionally remove detectable signals.

\noindent\textbf{Adversarial Attacks on AI Text Detectors.}
Recent work shows that AI text detectors are fragile under adversarial attacks. One line of work attacks detectors without using detector feedback, mainly through surface-level modifications. These methods reduce detector confidence via token-level or character-level perturbations, such as token substitutions and homoglyph replacements, while largely preserving the original text structure \citep{huang2024robust,creo2025silverspeak}.

% Another line of work uses LLMs to rewrite AI-generated content. General paraphrasing methods, such as direct rewriting and recursive paraphrasing, have been shown to weaken neural, zero-shot, and watermark-based detectors \citep{sadasivan2025can,krishna2023paraphrasing}. More adaptive methods further exploit detector feedback, either by constructing detector-guided supervised data\citep{krishna2023paraphrasing}, prompting LLMs to produce detector-evasive outputs\citep{lu2023large}, using detector scores for guided sampling\cite{cheng2026adversarial}, or applying reinforcement learning to optimize evasion\citep{nicks2024language,david2025authormist,ranganath2026stealthrl}. 
Another line of work uses LLMs to rewrite AI-generated content with detector feedback. 
\citet{sadasivan2025can} shows that direct rewriting and recursive paraphrasing by LLMs can substantially weaken neural, zero-shot, and watermark-based detectors. 
\citet{krishna2023paraphrasing} constructs paraphrase training data by using LLMs to generate candidate rewrite pairs and applying detectors to filter them, resulting in a supervised paraphraser that learns detector-evasive rewriting patterns. 
More adaptive methods incorporate detector signals directly into the attack process: \citet{lu2023large} uses detector feedback to construct prompts that induce detector-evasive generations; \citet{cheng2026adversarial} uses detector scores to guide candidate selection and iterative adversarial rewriting; and \citet{nicks2024language,david2025authormist,ranganath2026stealthrl} optimize generation policies with reinforcement learning, where detector-derived scores are used as rewards to encourage evasive outputs.

These RL fine-tuning approaches are related, but they usually combine detector evasion and semantic preservation into a single scalar objective. In contrast, we formulate detector-evasive generation as a constrained optimization problem, enforcing semantic preservation explicitly rather than treating it as a soft reward regularization.

\section{Problem Formulation}
% \vspace{-0.2cm}
\label{sec:problem_formulation}
Given an AI-generated input text $x \in \mathcal{X}$, the paraphrasing LLM, represented by a policy $\pi_\theta(y \mid x)$ parameterized by $\theta$, generates a rewritten text $y \in \mathcal{Y}$. The goal is to rewrite $x$ such that $y$ is less likely to be classified as AI-generated by a target detector, while preserving the semantics of the original text $x$. This setting naturally involves two possibly conflicting objectives: detector evasion and semantic preservation. Instead of linearly combining them into a single scalar reward, we treat semantic preservation as an explicit constraint.

\noindent\textbf{Detector-evasive paraphrasing.}
Denote by  $P_{\mathrm{detector}}(y)\in [0,1]$ the probability assigned by the detector that the rewritten text $y$ is AI-generated. We define the detector-evasion reward as
$
    r_{\mathrm{det}}(y) = 1 - P_{\mathrm{detector}}(y).
$
A higher value of $r_{\mathrm{det}}(y)$ indicates stronger detector evasion. The primary objective is to maximize the expected detector-evasion reward.

\noindent\textbf{Semantic preservation.}
Detector evasion alone is insufficient for paraphrasing tasks, since a model may reduce detector confidence by significantly changing the semantics of the original text, deleting important details, or producing unrelated content. Prior detector-evasion and paraphrase-attack work often measures semantic preservation with T5-based sentence-level similarity~\citep{ni2022sentencet5,ranganath2026stealthrl}. In contrast, we use BERTScore F1~\citep{zhang2020bertscore} as the semantic preservation metric, because its contextual token-level matching is better suited for capturing fine-grained semantic changes introduced during paraphrasing. We provide a sensitivity analysis and examples in Appendix~\ref{sec:fine_grained_semantic_sensitivity}. Specifically, define
\begin{equation}\label{eq:bert}
    r_{\mathrm{sem}}(x,y)
    =
    \mathrm{BERTScore}_{F1}(x,y),
\end{equation}
where $r_{\mathrm{sem}}(x,y) \in [0,1]$ measures the token-level semantic similarity between $x$ and $y$.

\noindent\textbf{CMDP formulation.}
We formulate detector-evasive paraphrasing as a Constrained Markov Decision Process (CMDP).  Let $\mathcal{D}$ denote the training dataset of AI-generated texts. The objective is to maximize detector evasion while satisfying the semantic preservation constraint:
\begin{align}\label{eq:cmdp_objective}
\max_{\theta} \ 
J_{\mathrm{det}}(\theta)
&=
\mathbb{E}_{x \sim \mathcal{D},\, y \sim \pi_\theta(\cdot \mid x)}
\left[
    r_{\mathrm{det}}(y)
\right]
\\
&\quad
-
\beta\,
\mathbb{D}_{\mathrm{KL}}
\left(
    \pi_\theta(\cdot \mid x)
    \,\middle\|\,
    \pi_{\mathrm{ref}}(\cdot \mid x)
\right),\nonumber
\\
\text{s.t.} \ 
J_{\mathrm{sem}}(\theta)
=&
\mathbb{E}_{x \sim \mathcal{D},\, y \sim \pi_\theta(\cdot \mid x)}
\left[
    r_{\mathrm{sem}}(x,y)
\right]
\ge
\tau_{\mathrm{sem}}.\nonumber
\end{align}
Here, $\pi_{\mathrm{ref}}$ is the reference policy and $\beta \geq 0$ controls the strength of the KL regularization, $\mathbb D_{\text{KL}}$  corresponds to the KL divergence.
We require the expected semantic similarity $J_{\mathrm{sem}}(\theta)$ to exceed a predefined threshold $\tau_{\mathrm{sem}}$. We will discuss the choice of $\tau_{\mathrm{sem}}$ later in \Cref{sec:thresholdtausem}.

\noindent\textbf{Lagrangian dual.}
To solve \cref{eq:cmdp_objective}, we introduce a dual variable $\lambda \geq 0$ and define the Lagrangian:
\begin{equation}
    \mathcal{L}(\theta,\lambda)
    =
    J_{\mathrm{det}}(\theta)
    +
    \lambda
    \left(
        J_{\mathrm{sem}}(\theta)
        -
        \tau_{\mathrm{sem}}
    \right).
\end{equation}
The original problem in \cref{eq:cmdp_objective} is equivalent to:
$
    \max_{\theta}
    \min_{\lambda \geq 0}
    \mathcal{L}(\theta,\lambda).
$
Its dual problem can be written as follows:
\begin{equation}
    \min_{\lambda \geq 0}
    \max_{\theta}
    \mathcal{L}(\theta,\lambda).
\end{equation}
We note that such a transformation incurs zero duality gap and is equivalent to the original problem in \cref{eq:cmdp_objective} \cite{altman1999constrained}. This min-max problem implies a dual-descent process: policy parameters $\theta$ are updated to maximize the Lagrangian, while the multiplier $\lambda$ is updated to minimize it, thereby penalizing the policy if the constraint is violated.

% Under this formulation, the policy parameters $\theta$ are updated to maximize detector evasion, while the dual variable $\lambda$ is adjusted according to the degree of semantic-constraint satisfaction. 
% When semantic similarity falls below the threshold, $\lambda$ increases and assigns greater weight to semantic preservation; when the constraint is satisfied, $\lambda$ decreases, allowing the policy to focus more on detector evasion. This yields a principled optimization framework in which detector evasion is improved only within the feasible region defined by semantic preservation.

\section{Detector Evasion Policy Optimization}
\label{sec:depo}

We introduce our algorithm DEPO, a constrained RL algorithm for detector-evasive paraphrasing. Unlike linear reward scalarization, DEPO adaptively adjusts the influence of the semantic constraint through a dual variable, allowing the policy to improve detector evasion only within the feasible semantic-preservation region.

To improve training efficiency and avoid the need for a separate value network, we introduce a group-based policy optimization procedure \cite{shao2024deepseekmath}. For each input text $x \sim \mathcal{D}$, we sample a group of $G$ candidate rewrites $\{y_i\}_{i=1}^{G}$ from the behavior policy $\pi_{\theta_{\mathrm{old}}}(\cdot \mid x)$. Each candidate rewrite is evaluated by two reward signals: a detector-evasion reward from the AI-text detector and a semantic-preservation reward based on BERTScore F1 as in \cref{eq:bert}.

\noindent\textbf{Reward Estimation.}
For each sampled rewrite $y_i$, we compute the detector-evasion reward as %\zou{fix this,  use the same notation as in eq. 1 $P_{detector}$}
$
    r_{\mathrm{det}}^{(i)}
    =
    1 - P_{\mathrm{detector}}(y_i).
$
To measure semantic preservation, we compute BERTScore F1 between the original input $x$ and the rewritten output $y_i$:
$
    r_{\mathrm{sem}}^{(i)}
    =
    \mathrm{BERTScore}_{F1}(x, y_i).
$
% In our CMDP formulation, $r_{\mathrm{det}}$ is optimized as the primary objective, whereas $r_{\mathrm{sem}}$ is used to enforce the semantic constraint
% \begin{equation}
%     \mathbb{E}_{\substack{x \sim \mathcal{D}\\ y \sim \pi_\theta(\cdot \mid x)}}
%     \left[
%         r_{\mathrm{sem}}(x,y)
%     \right]
%     \ge
%     \tau_{\mathrm{sem}},
% \end{equation}
% where $\tau_{\mathrm{sem}}$ is a predefined semantic-preservation threshold.

\noindent\textbf{Group-Relative Advantage Estimation.}
For each $x$, DEPO normalizes the detector and semantic rewards within the sampled group. Let
\begin{align}
    \mu_{\mathrm{det}}
    &=
    \frac{1}{G}
    \sum_{i=1}^{G}
    r_{\mathrm{det}}^{(i)},
    \\
    \sigma_{\mathrm{det}}
    &=
    \sqrt{
    \frac{1}{G}
    \sum_{i=1}^{G}
    \left(
        r_{\mathrm{det}}^{(i)}
        -
        \mu_{\mathrm{det}}
    \right)^2
    },
\end{align}
and similarly define $\mu_{\mathrm{sem}}$ and $\sigma_{\mathrm{sem}}$ for the semantic preservation rewards. The group-relative advantages are then computed as
\begin{align}
    A_{\mathrm{det}}^{(i)}
    =
    \frac{
        r_{\mathrm{det}}^{(i)}
        -
        \mu_{\mathrm{det}}
    }{
        \sigma_{\mathrm{det}}
    },
    A_{\mathrm{sem}}^{(i)}
    =
    \frac{
        r_{\mathrm{sem}}^{(i)}
        -
        \mu_{\mathrm{sem}}
    }{
        \sigma_{\mathrm{sem}}
    }.
\end{align}
This in-group normalization provides a prompt-specific baseline and reduces the variance of policy-gradient estimation.

\noindent\textbf{Primal-Dual Policy Optimization.}
At step $t$, DEPO maintains a Lagrange multiplier $\lambda_t \geq 0$ for the semantic preservation constraint. The combined advantage for the $i$-th rewrite is defined as
$
    A_{\mathrm{comb}}^{(i)}
    =
    A_{\mathrm{det}}^{(i)}
    +
    \lambda_t
    A_{\mathrm{sem}}^{(i)}.
$
The detector advantage drives the policy toward stronger detector evasion, while the semantic advantage is weighted by the current dual variable. When the semantic constraint is violated, $\lambda_t$ increases and gives greater importance to semantic preservation. When the constraint is satisfied, $\lambda_t$ decreases, allowing the policy to focus more on detector evasion.

\noindent\textit{Policy update.}
To update the policy parameters $\theta$, DEPO maximizes a clipped policy-gradient objective with a KL penalty against the frozen reference policy $\pi_{\mathrm{ref}}$. For each sampled rewrite $y_i$, define the importance sampling ratio
$
    \rho_i(\theta)
    =
    \frac{
        \pi_\theta(y_i \mid x)
    }{
        \pi_{\theta_{\mathrm{old}}}(y_i \mid x)
    }.
$
For a fixed input $x$, the policy optimization objective is
\begin{align}
\max_{\theta}& \mathcal{L}_{\mathrm{policy}}(\theta; x)
=\frac{1}{G}\sum_{i=1}^{G}\Big[\min\Big(\rho_i(\theta) A_{\mathrm{comb}}^{(i)}, \nonumber\\
&\operatorname{clip}\big(\rho_i(\theta),1-\epsilon,1+\epsilon\big)A_{\mathrm{comb}}^{(i)}
\Big) \nonumber\\
&-\beta D_{\mathrm{KL}}\big(\pi_\theta(y_i \mid x)\,\|\,\pi_{\mathrm{ref}}(y_i \mid x)
\big)
\Big],\label{eq:depo_policy_objective}
\end{align}
where $\epsilon$ is the clipping coefficient.% and $\beta$ controls the strength of the KL regularization. %The sequence-level log-ratio term serves as the sampled KL penalty, preventing the updated policy from drifting excessively from the reference model.

\noindent\textit{Dual update.}
After the policy update, DEPO updates the Lagrange multiplier according to the current degree of semantic-constraint satisfaction:
\begin{equation}
\begin{aligned}
    \lambda_{t+1}
    =
    \max
    \left(
        0,\,
        \lambda_t
        -
        \eta_\lambda
        \left(
            \mu_{\mathrm{sem}}
            -
            \tau_{\mathrm{sem}}
        \right)
    \right),\nonumber
\end{aligned}
\end{equation}
where $\eta_\lambda$ is the learning rate for the dual variable. If the average BERTScore falls below $\tau_{\mathrm{sem}}$, the term inside the parentheses becomes negative and $\lambda_t$ increases, thereby assigning greater weight to semantic preservation in subsequent policy updates. If the semantic constraint is satisfied, $\lambda_t$ decreases and reduces the influence of the constraint term.
% In practice, we may also update $\lambda_t$ using an exponential moving average of the semantic reward to reduce instability
% : \zou{fix this}
% \begin{equation}
% \begin{aligned}
%     \widehat{J}_{\mathrm{sem}}^{(t)}
%     =
%     \gamma
%     \widehat{J}_{\mathrm{sem}}^{(t-1)}
%     +
%     (1-\gamma)
%     \frac{1}{G}
%     \sum_{i=1}^{G}
%     r_{\mathrm{sem}}^{(i)} ,
%     \wang{exponential moving average}
% \end{aligned}
% \end{equation}
% and update the dual variable as
% \begin{equation}
% \begin{aligned}
%     \lambda_{t+1}
%     =
%     \max
%     \Big(
%         0,\,
%         \lambda_t
%         -
%         \eta_\lambda
%         \big(
%             \widehat{J}_{\mathrm{sem}}^{(t)}
%             -
%             \tau_{\mathrm{sem}}
%         \big)
%     \Big).
% \end{aligned}
% \end{equation}
% This smoothed update improves stability when reward estimates are noisy.
The complete algorithm is presented in \Cref{sec:pseudocode}. %DEPO alternates between sampling candidate rewrites, computing detector and semantic rewards, updating the policy through a constrained policy objective, and adapting the dual variable based on semantic-constraint satisfaction.

\section{Experiments}
\subsection{Experimental Setup}
\label{sec:experimental_setup}
%Given an AI-generated passage $x$, the attack LLM paraphrasing model generates a paraphrase $y$ that aims to reduce AI-text detector confidence while preserving the semantic content of $x$. This setting follows recent detector-evasion studies that evaluate whether rewriting-based attacks can make machine-generated text appear human-written without substantially changing the original meaning~\citep{david2025authormist,ranganath2026stealthrl}.

\noindent\textbf{Datasets and task.}
% Each input is an AI-generated passage $x$, and the task is to produce a semantically faithful paraphrase $y$ that is more likely to be classified as human-written. 
We use MAGE benchmark~\citep{li2024mage} as the primary training and in-domain evaluation corpus. To evaluate cross-domain transfer, we further test the trained paraphrasing policy on M4~\citep{wang2024m4}, RAID~\citep{dugan2024raid}, and the ICLR peer-review subset from \citet{yu2025peerreview}. In all evaluations, AI-written texts are used as inputs, while human-written texts are used for threshold calibration and AUC computation. Additional details of the peer-review dataset are provided in Appendix~\ref{sec:peer_review_dataset}. In \Cref{sec:cross_reward_validation}, we also train DEPO on a different RAID dataset using the RADAR detector for cross-reward validation.  

\noindent\textbf{Target detectors.}
During training, unless otherwise specified, DEPO uses the MAGE detector as the detector-reward model. At evaluation time, we test the attacked outputs against five detectors: MAGE detector~\citep{li2024mage}, OpenAI RoBERTa~\citep{solaiman2019release}, RADAR~\citep{hu2023radar}, Binoculars~\citep{hans2024spotting}, and Fast-DetectGPT~\citep{bao2024fastdetectgpt}. Details of these detectors are in \Cref{sec:baselineintro}. These detectors cover supervised neural classifiers, adversarially trained detectors, and zero-shot likelihood-based detectors. For the peer-review dataset, we additionally evaluate an in-domain RoBERTa detector trained on ICLR reviews.

\noindent\textbf{Training framework.}
All trainable variants are initialized from the same base paraphrasing model and fine-tuned with parameter-efficient LoRA adapters~\cite{hu2022lora}. We implement RL with the TRL library \cite{vonwerra2020trl}. Unless otherwise specified, the semantic threshold is set to $\tau_{\mathrm{sem}}=0.85$. Complete implementation details are provided in Table~\ref{tab:hyperparameters}. %Appendix~\ref{}.
Discussion on $\tau_{\mathrm{sem}}$ can be found in \Cref{sec:thresholdtausem}. We also conduct ablation studies with a different threshold $\tau_{\mathrm{sem}}=0.8, 0.85, 0.9$ (see \Cref{sec:ablation_tau_sem}).

\noindent\textbf{Evaluation metrics.}
We report detector reward, semantic reward, constraint fulfillment, attack success rate (ASR), $\Delta$AUC, and output length. Detailed definitions of these metrics can be found in \Cref{sec:baselineintro}. All detector scores are oriented so that larger scores indicate stronger confidence that the input is AI-generated.

\subsection{Evasion vs. Semantics}

Table~\ref{tab:mage_detector_combined_eval} presents the results against the MAGE detector. The attack policy is trained with MAGE detector feedback. We compare DEPO with existing paraphrase attacks, single-objective ablations, and reward linearization baselines. Details of these baselines are in \Cref{sec:baselineintro}.

\definecolor{rowgray}{gray}{0.95}
\begin{table*}[!htb]
\centering
\small
\setlength{\tabcolsep}{3pt}
\renewcommand{\arraystretch}{1.15}
\caption{Main detector-evasion results against the MAGE detector on the evaluation set. The Original row reports detector performance on unmodified AI-written texts before attack. Detector reward, ASR, and ASR@1\%FPR measure attack effectiveness; semantic reward and constraint fulfillment measure semantic preservation. The semantic threshold is $\tau_{\mathrm{sem}}=0.85$.}
\label{tab:mage_detector_combined_eval}
% \resizebox{\textwidth}{!}{
{
\begin{tabular}{lcccccc}
\toprule
\textbf{Method}
& \makecell{\textbf{Detector}\\\textbf{Reward} $\uparrow$}
& \makecell{\textbf{Semantic}\\\textbf{Reward} $\uparrow$}
& \makecell{\textbf{Constraint}\\\textbf{Fulfillment}}
& \textbf{ASR@$\tau=0.5$ $\uparrow$}
& \textbf{ASR@1\%FPR $\uparrow$}
& \textbf{$\Delta$AUC} $\downarrow$ \\
% & \makecell{\textbf{Length}\\\textbf{of Text}} \\
\midrule

\rowcolor{rowgray}
Original
& --
& --
& --
& $0.007$
& $0.078$
& $0.494$ \\
% & $1118$ \\

\midrule

Direct Paraphrasing
& $0.040$
& $0.854$
& \cmark
& $0.030$
& $0.276$
& $0.476$ \\
% & $974$ \\

AuthorMist
& $0.070$
& $0.810$
& \xmark
& $0.059$
& $0.321$
& $0.466$ \\
% & $899$ \\

StealthRL
& $0.045$
& $0.613$
& \xmark
& $0.021$
& $0.789$
& $0.459$ \\
% & $301$ \\

SilverSpeak
& $0.028$
& $0.338$
& \xmark
& $0.005$
& $0.849$
& $0.469$ \\
% & $1118$ \\

Adv. Paraphrase 
& $0.073$ 
& $0.826$ 
& \xmark 
& $0.094$ 
& $0.489$ 
& $0.453$ \\
% & $930$ \\

\midrule

BERTScore 0 (Detector Evasion Only)
& $0.903$
& $0.798$
& \xmark
& $0.895$
& $0.995$
& $0.028$ \\
% & $1001$ \\

BERTScore 1 (Semantic Only)
& $0.010$
& $0.993$
& \cmark
& $0.008$
& $0.081$
& $0.494$ \\
% & $1108$ \\

BERTScore Linear 0.1
& $0.883$
& $0.805$
& \xmark
& $0.913$
& $0.994$
& $0.050$ \\
% & $1012$ \\

BERTScore Linear 0.3
& $0.758$
& $0.834$
& \xmark
& $0.768$
& $0.953$
& $0.081$ \\
% & $1004$ \\

BERTScore Linear 0.5
& $0.348$
& $0.837$
& \xmark
& $0.347$
& $0.668$
& $0.344$ \\
% & $1188$ \\

\midrule

\rowcolor{rowgray}
\textbf{DEPO}
& $\mathbf{0.716}$
& $\mathbf{0.857}$
& \cmark
& $\mathbf{0.725}$
& $\mathbf{0.957}$
& $\mathbf{0.140}$ \\
% & $\mathbf{1025}$ \\

\bottomrule
\end{tabular}
}
\end{table*}

\noindent\textbf{Results.}
Table~\ref{tab:mage_detector_combined_eval} shows a clear trade-off between detector evasion and semantic preservation. Existing paraphrase attacks either provide limited evasion or sacrifice semantic fidelity. For example, Direct Paraphrasing satisfies the semantic constraint with a BERTScore of $0.854$, but only reaches an ASR of $0.030$, while stronger baselines such as AuthorMist and Adversarial Paraphrase improve ASR but fall below the semantic preservation threshold.

Optimizing detector evasion alone achieves the strongest attack, with ASR $0.895$ and AUC $0.472$, but violates the semantic preservation constraint. In contrast, optimizing semantic preservation alone almost perfectly preserves meaning but produces little evasion. Fixed linear scalarization interpolates between these extremes, but its performance is highly sensitive to the weights and does not precisely preserve semantics.

Our DEPO achieves the best trade-off. Among methods satisfying the semantic preservation constraint of $\tau_{\mathrm{sem}}=0.85$, DEPO obtains the strongest evasion, reaching ASR $0.725$ and $\Delta$AUC $0.140$ while maintaining a BERTScore of $0.857$. These results match our CMDP formulation:  optimize detector evasion while remaining within the prescribed semantic preservation.

%Compared with recent detector-evasion baselines, DEPO provides a stronger and more reliable attack under semantic constraints. Direct Paraphrasing preserves meaning but produces only limited evasion, while AuthorMist, StealthRL, SilverSpeak, and Adversarial Paraphrase improve detector evasion to different extents at the cost of lower semantic fidelity. In contrast, DEPO is the only optimized attack that both satisfies the semantic preservation threshold and achieves strong attack effectiveness, reaching ASR $0.725$ and ASR@1\%FPR $0.957$ with a BERTScore of $0.857$. This suggests that DEPO does not achieve evasion solely through deletion, surface corruption, or semantic drift; instead, it learns a constrained paraphrasing policy that substantially weakens detector performance while preserving the original semantics.

\subsection{Cross Detector Generalization}
\label{sec:detector_wise_analysis}

Table~\ref{tab:attack_mage} evaluates the attacked outputs from the MAGE evaluation set under different AI-text detectors. This setting tests whether a policy trained with MAGE detector feedback can also compromise other detectors on the same dataset.
%ASR is computed using the threshold protocol defined in Section~\ref{sec:experimental_setup}: supervised probability-based detectors use $\tau=0.5$, while zero-shot detectors use validation-calibrated thresholds.

% ===== MAGE =====
\begin{table*}[!htbp]
\centering
\small
\setlength{\tabcolsep}{2pt}
\renewcommand{\arraystretch}{1.15}
\caption{Same-domain cross-detector evaluation on the MAGE evaluation set. The attack policy is trained with MAGE detector feedback and evaluated against different supervised and zero-shot detectors.}
\label{tab:attack_mage}
{
\begin{tabular}{lc cc cc cc cc cc}
\toprule
\multirow{2}{*}{\textbf{Method}}
& \multirow{2}{*}{\makecell{\textbf{BERT}\\\textbf{Score}}}
& \multicolumn{2}{c}{\textbf{MAGE}}
& \multicolumn{2}{c}{\textbf{RoBERTa}}
& \multicolumn{2}{c}{\textbf{RADAR}}
& \multicolumn{2}{c}{\textbf{Binoculars}}
& \multicolumn{2}{c}{\textbf{Fast-DetectGPT}} \\
\cmidrule(lr){3-4}
\cmidrule(lr){5-6}
\cmidrule(lr){7-8}
\cmidrule(lr){9-10}
\cmidrule(lr){11-12}
& 
& \textbf{ASR$\uparrow$} & \textbf{$\Delta$AUC$\downarrow$}
& \textbf{ASR$\uparrow$} & \textbf{$\Delta$AUC$\downarrow$}
& \textbf{ASR$\uparrow$} & \textbf{$\Delta$AUC$\downarrow$}
& \textbf{ASR$\uparrow$} & \textbf{$\Delta$AUC$\downarrow$}
& \textbf{ASR$\uparrow$} & \textbf{$\Delta$AUC$\downarrow$} \\
\midrule
\rowcolor{rowgray}
Original
& --
& $0.01$ & $0.49$ & $0.32$ & $0.32$ & $0.28$ & $0.22$ & $0.47$ & $0.19$ & $0.53$ & $0.15$ \\

\midrule
Direct Paraphrasing
& $0.854$
& $0.03$ & $0.48$ & $0.47$ & $0.23$ & $0.26$ & $0.25$ & $0.74$ & $0.09$ & $0.74$ & $0.10$ \\

AuthorMist
& $0.810$
& $0.06$ & $0.47$ & $0.57$ & $0.15$ & $0.35$ & $0.19$ & $0.95$ & $0.17$ & $0.96$ & $0.16$ \\

StealthRL
& $0.613$
& $0.02$ & $0.46$ & $0.54$ & $0.21$ & $0.01$ & $0.48$ & $0.99$ & $0.45$ & $1.00$ & $0.41$ \\

SilverSpeak
& $0.338$
& $0.01$ & $0.47$ & $0.00$ & $0.48$ & $0.00$ & $0.33$ & $0.20$ & $0.46$ & $0.99$ & $0.16$ \\

Adv. Paraphrase
& $0.826$
& $0.09$ & $0.45$ & $0.56$ & $0.20$ & $0.29$ & $0.26$ & $0.82$ & $0.04$ & $0.81$ & $0.06$ \\

\midrule
MAGE BERTScore 0
& $0.798$
& $0.91$ & $0.05$ & $0.63$ & $0.10$ & $0.44$ & $0.14$ & $0.98$ & $0.23$ & $0.97$ & $0.20$ \\

MAGE BERTScore Linear 0.1
& $0.805$
& $0.90$ & $0.03$ & $0.61$ & $0.12$ & $0.26$ & $0.25$ & $0.95$ & $0.09$ & $0.94$ & $0.09$ \\

MAGE BERTScore Linear 0.3
& $0.834$
& $0.77$ & $0.08$ & $0.61$ & $0.11$ & $0.45$ & $0.13$ & $0.95$ & $0.18$ & $0.95$ & $0.14$ \\

\dashedmidrule

MAGE BERTScore 1
& $0.993$
& $0.01$ & $0.49$ & $0.31$ & $0.32$ & $0.28$ & $0.23$ & $0.47$ & $0.20$ & $0.53$ & $0.15$ \\

MAGE BERTScore Linear 0.5
& $0.837$
& $0.35$ & $0.34$ & $0.43$ & $0.26$ & $0.10$ & $0.36$ & $0.81$ & $\mathbf{0.07}$ & $0.80$ & $\mathbf{0.03}$ \\

\rowcolor{rowgray}
\textbf{DEPO}
& $0.857$
& $\mathbf{0.73}$ & $\mathbf{0.14}$ & $\mathbf{0.58}$ & $\mathbf{0.14}$ & $\mathbf{0.36}$ & $\mathbf{0.19}$ & $\mathbf{0.92}$ & $0.12$ & $\mathbf{0.91}$ & $0.09$ \\

\bottomrule
\end{tabular}
}
\end{table*}

\noindent\textbf{Results.}
Compared with the original texts, DEPO consistently increases ASR and lowers AUC across all five detectors, including supervised detectors and zero-shot likelihood-based detectors. The effect is strongest on the reward detector, where DEPO increases MAGE ASR from $0.01$ to $0.73$ while maintaining a BERTScore of $0.857$. It also transfers to unseen detectors, achieving ASR scores of $0.58$ on RoBERTa, $0.36$ on RADAR, $0.92$ on Binoculars, and $0.91$ on Fast-DetectGPT.
Some of the other baselines also generalize across different detectors, but a similar observation can be made: they fail to precisely control semantic preservation.

% Unconstrained or low-semantic-weight variants sometimes achieve stronger evasion, but they do so with substantially lower semantic preservation. Among constraint satisfying methods, DEPO provides the strongest overall detector evasion: it substantially improves over semantic-only optimization and outperforms high-similarity paraphrasing baselines across both supervised and zero-shot detectors. These results suggest that DEPO learns detector-evasive paraphrasing patterns that generalize beyond the MAGE detector while preserving the intended meaning of the original text.

% Compared with recent paraphrase-evasion baselines, DEPO offers a more balanced and reliable attack profile. Direct Paraphrasing preserves meaning but yields weak evasion on the MAGE detector, while AuthorMist, StealthRL, SilverSpeak, and Adversarial Paraphrase often improve evasion by sacrificing semantic fidelity. DEPO is the only optimized attack in this comparison that maintains the semantic constraint and achieves strong transfer across all evaluated detectors. This supports the effectiveness of constrained policy optimization for detector-evasive paraphrasing.

% \vspace{-0.15cm}
\subsection{Cross Domain Generalization}
\label{sec:cross_domain_generalization}
% \vspace{-0.1cm}

Table~\ref{tab:attack_m4} (and \Cref{tab:attack_reviews,tab:attack_raid} in the Appendix) evaluates whether the MAGE-trained policy transfers beyond the training domain. We keep the attack model fixed and test its outputs on three held-out corpora using the same detector suite:
\begin{itemize}[leftmargin=*, itemsep=1pt, topsep=2pt, parsep=0pt, partopsep=0pt]
    \item \textbf{M4}~\citep{wang2024m4}: a multi-generator, multi-domain, and multilingual benchmark for machine-generated text detection.
    \item \textbf{RAID}~\citep{dugan2024raid}: a robustness benchmark covering diverse generators, domains, decoding strategies, and adversarial attacks.
    \item \textbf{Peer review}~\citep{yu2025peerreview}: an academic peer-review dataset containing human and AI-written ICLR reviews from 2017--2022. Details are in Appendix~\ref{sec:peer_review_dataset}.
\end{itemize}
For each dataset, we use the evaluation protocol described in Section~\ref{sec:experimental_setup}.
% The in-domain peer-review detector is validated in Appendix~\ref{sec:peer_review_detector_validation}.

% ===== M4 =====
\begin{table*}[htbp]
\centering
\small
\setlength{\tabcolsep}{2pt}
\renewcommand{\arraystretch}{1.15}
\caption{Attack effectiveness on the M4 evaluation set under different AI-text detectors.}
\label{tab:attack_m4}
% \resizebox{\textwidth}{!}{
{
\begin{tabular}{lc cc cc cc cc cc}
\toprule
\multirow{2}{*}{\textbf{Method}}
& \multirow{2}{*}{\makecell{\textbf{BERT}\\\textbf{Score}}}
& \multicolumn{2}{c}{\textbf{MAGE}}
& \multicolumn{2}{c}{\textbf{RoBERTa}}
& \multicolumn{2}{c}{\textbf{RADAR}}
& \multicolumn{2}{c}{\textbf{Binoculars}}
& \multicolumn{2}{c}{\textbf{Fast-DetectGPT}} \\
\cmidrule(lr){3-4}
\cmidrule(lr){5-6}
\cmidrule(lr){7-8}
\cmidrule(lr){9-10}
\cmidrule(lr){11-12}
&
& \textbf{ASR$\uparrow$} & \textbf{$\Delta$AUC$\downarrow$}
& \textbf{ASR$\uparrow$} & \textbf{$\Delta$AUC$\downarrow$}
& \textbf{ASR$\uparrow$} & \textbf{$\Delta$AUC$\downarrow$}
& \textbf{ASR$\uparrow$} & \textbf{$\Delta$AUC$\downarrow$}
& \textbf{ASR$\uparrow$} & \textbf{$\Delta$AUC$\downarrow$} \\
\midrule
\rowcolor{rowgray}
Original
& -- 
& $0.02$ & $0.47$
& $0.84$ & $0.03$
& $0.36$ & $0.14$
& $0.93$ & $0.08$
& $0.74$ & $0.01$ \\

\midrule
Direct Paraphrasing
& $0.856$
& $0.06$ & $0.41$
& $0.77$ & $0.04$
& $0.38$ & $0.13$
& $0.93$ & $0.01$
& $0.77$ & $0.07$ \\

AuthorMist
& $0.817$
& $0.04$ & $0.42$
& $0.82$ & $0.00$
& $0.48$ & $0.05$
& $0.98$ & $0.26$
& $0.97$ & $0.23$ \\

StealthRL
& $0.656$
& $0.01$ & $0.34$
& $0.77$ & $0.15$
& $0.02$ & $0.44$
& $1.00$ & $0.43$
& $0.99$ & $0.41$ \\

SilverSpeak
& $0.391$
& $0.03$ & $0.35$
& $0.00$ & $0.48$
& $0.00$ & $0.30$
& $0.66$ & $0.45$
& $0.99$ & $0.28$ \\

Adv. Paraphrase
& $0.856$
& $0.13$ & $0.39$
& $0.88$ & $0.01$
& $0.73$ & $0.02$
& $0.80$ & $0.01$
& $0.79$ & $0.09$ \\

\midrule
MAGE BERTScore 0
& $0.804$
& $0.85$ & $0.18$
& $0.85$ & $0.03$
& $0.64$ & $0.04$
& $0.98$ & $0.27$
& $0.96$ & $0.24$ \\

MAGE BERTScore Linear 0.1
& $0.791$
& $0.90$ & $0.21$
& $0.86$ & $0.06$
& $0.39$ & $0.13$
& $0.97$ & $0.15$
& $0.95$ & $0.15$ \\

MAGE BERTScore Linear 0.3
& $0.839$
& $0.68$ & $0.04$
& $0.85$ & $0.04$
& $0.57$ & $0.01$
& $0.96$ & $0.21$
& $0.92$ & $0.18$ \\

\dashedmidrule

MAGE BERTScore 1
& $0.990$
& $0.02$ & $0.46$
& $0.83$ & $\mathbf{0.00}$
& $0.36$ & $0.14$
& $0.93$ & $0.06$
& $0.73$ & $\mathbf{0.03}$ \\

MAGE BERTScore Linear 0.5
& $0.827$
& $0.29$ & $0.23$
& $0.73$ & $0.09$
& $0.18$ & $0.29$
& $\mathbf{0.97}$ & $\mathbf{0.03}$
& $0.87$ & $0.04$ \\

\rowcolor{rowgray}
\textbf{DEPO}
& $0.854$
& $\mathbf{0.70}$ & $\mathbf{0.04}$
& $\mathbf{0.83}$ & ${0.02}$
& $\mathbf{0.51}$ & $\mathbf{0.04}$
& $0.95$ & $0.22$
& $\mathbf{0.93}$ & ${0.17}$ \\

\bottomrule
\end{tabular}
}
\end{table*}

% ===== RAID =====

\noindent\textbf{Peer-review RoBERTa detector.}
For the peer-review evaluation, we additionally train a RoBERTa-based detector on ICLR 2017--2019 reviews from the AI Peer Review Detection Benchmark \citep{yu2025peerreview}. The detector is evaluated on held-out ICLR 2020--2022 reviews, which tests temporal generalization within the same peer-review domain. As shown in Table~\ref{tab:peer_review_detector_sota} (in the Appendix), this detector achieves very high TPR under low-FPR operating points, outperforming the strongest reported baseline detector, Anchor~\cite{yu2025peerreview}, from the original benchmark.

\noindent\textbf{Results.}
\Cref{tab:attack_m4} (and \Cref{tab:attack_raid,tab:attack_reviews} in the Appendix) show that DEPO generalizes beyond the MAGE training domain. On M4, DEPO maintains strong attack performance against several detectors, reaching ASR $0.70$ on MAGE, $0.51$ on RADAR, $0.95$ on Binoculars, and $0.93$ on Fast-DetectGPT, while keeping a BERTScore of $0.854$, satisfying the prespecified threshold. Because RoBERTa is already weak on the original M4 texts, DEPO adds little but remains competitive with the strongest semantically faithful baselines.

The transfer effect is clearer on RAID. Although RAID differs from MAGE in generators, domains, and attack settings, DEPO reduces detector separability across all five detectors, including the target-family MAGE detector and unseen detectors. It reaches ASR $0.86$ on MAGE and $0.99$ on RoBERTa, and also improves evasion against RADAR, Binoculars, and Fast-DetectGPT while maintaining a BERTScore of $0.878$. Some unconstrained variants achieve stronger evasion, but at the cost of lower semantic similarity. DEPO instead provides a more balanced cross-detector profile under semantic control.

On peer-review data, DEPO remains effective for general-purpose detectors, with ASR $0.94$ on MAGE, $0.84$ on RADAR, $0.95$ on Binoculars, and $0.78$ on Fast-DetectGPT. The in-domain PeerReview RoBERTa detector shows lower AUC degradation, suggesting that a detector trained directly on peer-review data captures domain-specific signals that transfer attacks do not fully remove. Nevertheless, DEPO increases threshold-based evasion on this detector while preserving semantics more than most attack baselines. 

Overall, DEPO learns rewriting patterns that transfer across domains and detector families.

% \vspace{-0.15cm}
\subsection{Cross-Reward Validation}
\label{sec:cross_reward_validation}
% \vspace{-0.1cm}

To test whether DEPO generalizes beyond a single training corpus and reward detector, we conduct a cross-reward validation. Specifically, we train a separate DEPO policy on RAID using RADAR as a detector. This setting changes both the training distribution and the reward detector.

\definecolor{rowgray}{gray}{0.95}
\begin{table*}[h]
\centering
\small
\setlength{\tabcolsep}{3pt}
\renewcommand{\arraystretch}{1.15}
\caption{Main results of DEPO trained on RAID with RADAR detector feedback. The Original row reports RADAR detector performance on unmodified AI-written texts. The semantic threshold is $\tau_{\mathrm{sem}}=0.85$.}
\label{tab:radar_detector_raid_combined_eval}
% \resizebox{\textwidth}{!}{
{
\begin{tabular}{lcccccc}
\toprule
\textbf{Method}
& \makecell{\textbf{Detector}\\\textbf{Reward} $\uparrow$}
& \makecell{\textbf{Semantic}\\\textbf{Reward} $\uparrow$}
& \makecell{\textbf{Constraint}\\\textbf{Fulfillment}}
& \textbf{ASR@$\tau=0.5$ $\uparrow$}
& \textbf{ASR@1\%FPR $\uparrow$}
& \textbf{$\Delta$AUC} $\downarrow$ \\
% & \makecell{\textbf{Length}\\\textbf{of Text}} \\
\midrule

\rowcolor{rowgray}
Original
& --
& --
& --
& $0.444$
& $0.631$
& $0.419$ \\
% & $1910$ \\

\midrule

Direct Paraphrasing
& $0.652$
& $0.877$
& \cmark
& $0.720$
& $0.835$
& $0.366$ \\
% & $1416$ \\

AuthorMist
& $0.789$
& $0.825$
& \xmark
& $0.839$
& $0.891$
& $0.240$ \\
% & $1273$ \\

StealthRL
& $0.259$
& $0.733$
& \xmark
& $0.256$
& $0.337$
& $0.460$ \\
% & $626$ \\

SilverSpeak
& $0.327$
& $0.374$
& \xmark
& $0.136$
& $0.634$
& $0.486$ \\
% & $1910$ \\

Adv. Paraphrase 
& $0.697$ 
& $0.876$ 
& \cmark 
& $0.770$ 
& $0.865$ 
& $0.348$ \\
% & $1407$ \\

\midrule

BERTScore 0 (Detector Evasion Only)
& $0.973$
& $0.775$
& \xmark
& $0.983$
& $0.986$
& $0.287$ \\
% & $1419$ \\

BERTScore 1 (Semantic Only)
& $0.475$
& $0.991$
& \cmark
& $0.458$
& $0.653$
& $0.413$ \\
% & $1891$ \\

BERTScore Linear 0.1
& $0.961$
& $0.783$
& \xmark
& $0.971$
& $0.980$
& $0.274$ \\
% & $1390$ \\

\midrule

\rowcolor{rowgray}
\textbf{DEPO}
& $\mathbf{0.931}$
& $\mathbf{0.850}$
& \cmark
& $\mathbf{0.952}$
& $\mathbf{0.964}$
& $\mathbf{0.095}$ \\
% & $\mathbf{1474}$ \\

\bottomrule
\end{tabular}
}
\end{table*}

\noindent\textbf{Target-detector results.}
Table~\ref{tab:radar_detector_raid_combined_eval} shows that DEPO remains effective when trained on RAID with RADAR feedback. Compared with Direct Paraphrasing and Adversarial Paraphrase, DEPO achieves a much stronger attack against RADAR while still satisfying the semantic constraint. It reaches ASR $0.952$ and $\Delta$AUC $0.095$ with a BERTScore of $0.850$. Detector-only optimization obtains lower AUC, but its BERTScore drops to $0.775$, indicating that its attack strength comes with clear semantic degradation. These results confirm that DEPO can optimize a different detector-evasion objective while keeping semantic preservation under explicit control.

\noindent\textbf{Comparison with MAGE-trained DEPO.}
Table~\ref{tab:attack_raid_radar_trained} (in the Appendix) further shows that the reward detector strongly shapes the learned attack behavior. Compared with the MAGE-trained DEPO policy evaluated on RAID in Table~\ref{tab:attack_raid}, the RADAR-trained policy is much stronger against RADAR, increasing ASR from $0.65$ to $0.95$ and lowering $\Delta$AUC from $0.32$ to $0.09$. In contrast, its attack against the MAGE detector becomes weaker, with ASR decreasing from $0.86$ to $0.19$. This contrast is expected: each policy is optimized against a different reward detector. The RADAR-trained policy also transfers to several unseen detectors, reaching ASR $0.98$ on RoBERTa, $0.87$ on Binoculars, and $0.84$ on Fast-DetectGPT. Overall, the results show that DEPO is not tied to MAGE or to a single dataset, but the strongest attack direction is determined primarily by the detector used for reward feedback.

AUC-level comparisons between MAGE-trained and RADAR-trained policies are in Appendix~\ref{sec:reward_detector_auc_analysis}.

Together, these results provide a cross-validation of DEPO: the same constrained optimization framework remains effective when both the training dataset and reward detector are changed, while the detector used for reward feedback determines where the learned attack is strongest.

% \vspace{-0.25cm}
\subsection{Summary}
% \vspace{-0.2cm}
The results show that DEPO improves detector evasion while keeping semantic preservation under explicit control. Its gains are not confined to the reward detector or the training corpus, and the cross-reward experiment confirms that the framework remains effective when both the training data and detector feedback are changed. These findings support the central claim that constrained policy optimization provides a more reliable way to control the evasion--semantics trade-off than unconstrained optimization or fixed reward scalarization. 
% Additional prompt-robustness and score-distribution analyses are provided in Appendix~\ref{sec:prompt_robustness} and Appendix~\ref{sec:score_distribution_analysis}.
Additionally, to further examine DEPO's generalization behavior, prompt stability, and output quality, we analyze detection score distribution change under attacking, reward detector effects on AUC, prompt robustness, and quality evaluation of paraphrased texts in Appendices~\ref{sec:score_distribution_analysis}, ~\ref{sec:reward_detector_auc_analysis}, ~\ref{sec:prompt_robustness}, and ~\ref{sec:llm_judge}.
% An interactive demo of the DEPO model is also available at \url{https://anonymous-submission-demo-paraphrase-demo.hf.space}, where readers can try paraphrasing. %anonymous-submission
An interactive demo of the DEPO model is also available at \url{https://huggingface.co/spaces/WizardWang01/depo-paraphrase-demo}, where readers can try paraphrasing. % open-source

% \vspace{-0.1cm}
\section{Conclusion}
% \vspace{-0.1cm}

We present DEPO, a constrained RL-based framework for optimizing AI-generated text to evade detection while preserving semantic meaning and generation quality. DEPO formulates detector evasion as a constrained RL problem, optimizing detector evasion while preserving semantics. This design enables DEPO to move beyond surface-level paraphrasing and learn adaptive rewriting policies that remain faithful to the original content.
Experiments on various datasets show that DEPO achieves strong detector evasion within the prescribed semantic-preservation region. Cross-detector, cross-domain, and cross-reward evaluations further indicate that DEPO is not limited to a single detector or dataset, while the reward detector determines where the learned attack is strongest. %Additional analysis shows that BERTScore is better suited than sentence-level T5 similarity for capturing fine-grained structural and logic-sensitive semantic changes.
We also note that BERTScore is not a complete semantic-equivalence measure, but it provides a more local, token-sensitive constraint than sentence-level embedding similarity for the perturbations considered here.

\newpage
\section*{Limitations}

DEPO has dual-use implications. We study detector-evasive paraphrasing to stress-test detector robustness and motivate stronger defenses, but similar techniques could be misused to conceal AI-generated text. Future work should combine constrained attack evaluation with human assessment, stronger semantic constraints, and adaptive defense mechanisms.

\bibliography{custom}

@article{perkins2023game,
  title={Game of Tones: Faculty detection of GPT-4 generated content in university assessments},
  author={Perkins, Mike and Roe, Jasper and Postma, Darius and McGaughran, James and Hickerson, Don},
  journal={arXiv preprint arXiv:2305.18081},
  year={2023}
}

@article{li2024faster,
  title={Faster algorithm and sharper analysis for constrained Markov decision process},
  author={Li, Tianjiao and Guan, Ziwei and Zou, Shaofeng and Xu, Tengyu and Liang, Yingbin and Lan, Guanghui},
  journal={Operations Research Letters},
  volume={54},
  pages={107107},
  year={2024},
  publisher={Elsevier}
}

@inproceedings{hanley2024machine,
  title={Machine-made media: Monitoring the mobilization of machine-generated articles on misinformation and mainstream news websites},
  author={Hanley, Hans WA and Durumeric, Zakir},
  booktitle={Proceedings of the international AAAI conference on web and social media},
  volume={18},
  pages={542--556},
  year={2024}
}

@article{zhang2020bertscore,
  title={Bertscore: Evaluating text generation with bert},
  author={Zhang, Tianyi and Kishore, Varsha and Wu, Felix and Weinberger, Kilian Q and Artzi, Yoav},
  journal={arXiv preprint arXiv:1904.09675},
  year={2019}
}

@inproceedings{ni2022sentencet5,
  title={Sentence-t5: Scalable sentence encoders from pre-trained text-to-text models},
  author={Ni, Jianmo and Abrego, Gustavo Hernandez and Constant, Noah and Ma, Ji and Hall, Keith and Cer, Daniel and Yang, Yinfei},
  booktitle={Findings of the association for computational linguistics: ACL 2022},
  pages={1864--1874},
  year={2022}
}

@inproceedings{mahajan2024alignsim,
  title={ALIGN-SIM: A task-free test bed for evaluating and interpreting sentence embeddings through semantic similarity alignment},
  author={Mahajan, Yash and Bansal, Naman and Blanco, Eduardo and Karmaker, Santu},
  booktitle={Findings of the Association for Computational Linguistics: EMNLP 2024},
  pages={7393--7428},
  year={2024}
}

@inproceedings{gehrmann2019gltr,
  title={Gltr: Statistical detection and visualization of generated text},
  author={Gehrmann, Sebastian and Strobelt, Hendrik and Rush, Alexander M},
  booktitle={Proceedings of the 57th annual meeting of the association for computational linguistics: system demonstrations},
  pages={111--116},
  year={2019}
}

@article{solaiman2019release,
  title={Release strategies and the social impacts of language models},
  author={Solaiman, Irene and Brundage, Miles and Clark, Jack and Askell, Amanda and Herbert-Voss, Ariel and Wu, Jeff and Radford, Alec and Krueger, Gretchen and Kim, Jong Wook and Kreps, Sarah and others},
  journal={arXiv preprint arXiv:1908.09203},
  year={2019}
}

@inproceedings{ippolito2020automatic,
  title={Automatic detection of generated text is easiest when humans are fooled},
  author={Ippolito, Daphne and Duckworth, Daniel and Callison-Burch, Chris and Eck, Douglas},
  booktitle={Proceedings of the 58th annual meeting of the association for computational linguistics},
  pages={1808--1822},
  year={2020}
}

@inproceedings{uchendu2021turingbench,
  title={Turingbench: A benchmark environment for turing test in the age of neural text generation},
  author={Uchendu, Adaku and Ma, Zeyu and Le, Thai and Zhang, Rui and Lee, Dongwon},
  booktitle={Findings of the association for computational linguistics: EMNLP 2021},
  pages={2001--2016},
  year={2021}
}

@inproceedings{li2024mage,
  title={MAGE: Machine-generated text detection in the wild},
  author={Li, Yafu and Li, Qintong and Cui, Leyang and Bi, Wei and Wang, Zhilin and Wang, Longyue and Yang, Linyi and Shi, Shuming and Zhang, Yue},
  booktitle={Proceedings of the 62nd Annual Meeting of the Association for Computational Linguistics (Volume 1: Long Papers)},
  pages={36--53},
  year={2024}
}

@inproceedings{wang2024m4,
  title={M4: Multi-generator, multi-domain, and multi-lingual black-box machine-generated text detection},
  author={Wang, Yuxia and Mansurov, Jonibek and Ivanov, Petar and Su, Jinyan and Shelmanov, Artem and Tsvigun, Akim and Whitehouse, Chenxi and Afzal, Osama Mohammed and Mahmoud, Tarek and Sasaki, Toru and others},
  booktitle={Proceedings of the 18th Conference of the European Chapter of the Association for Computational Linguistics (Volume 1: Long Papers)},
  pages={1369--1407},
  year={2024}
}

@inproceedings{dugan2024raid,
  title={Raid: A shared benchmark for robust evaluation of machine-generated text detectors},
  author={Dugan, Liam and Hwang, Alyssa and Trhl{\'\i}k, Filip and Zhu, Andrew and Ludan, Josh Magnus and Xu, Hainiu and Ippolito, Daphne and Callison-Burch, Chris},
  booktitle={Proceedings of the 62nd Annual Meeting of the Association for Computational Linguistics (Volume 1: Long Papers)},
  pages={12463--12492},
  year={2024}
}

@article{zellers2019defending,
  title={Defending against neural fake news},
  author={Zellers, Rowan and Holtzman, Ari and Rashkin, Hannah and Bisk, Yonatan and Farhadi, Ali and Roesner, Franziska and Choi, Yejin},
  journal={Advances in neural information processing systems},
  volume={32},
  year={2019}
}

@inproceedings{verma2024ghostbuster,
  title={Ghostbuster: Detecting text ghostwritten by large language models},
  author={Verma, Vivek and Fleisig, Eve and Tomlin, Nicholas and Klein, Dan},
  booktitle={Proceedings of the 2024 Conference of the North American Chapter of the Association for Computational Linguistics: Human Language Technologies (Volume 1: Long Papers)},
  pages={1702--1717},
  year={2024}
}

@inproceedings{mitchell2023detectgpt,
  title={Detectgpt: Zero-shot machine-generated text detection using probability curvature},
  author={Mitchell, Eric and Lee, Yoonho and Khazatsky, Alexander and Manning, Christopher D and Finn, Chelsea},
  booktitle={International conference on machine learning},
  pages={24950--24962},
  year={2023},
  organization={PMLR}
}

@inproceedings{bao2024fastdetectgpt,
  title={Fast-detectgpt: Efficient zero-shot detection of machine-generated text via conditional probability curvature},
  author={Bao, Guangsheng and Zhao, Yanbin and Teng, Zhiyang and Yang, Linyi and Zhang, Yue},
  booktitle={International Conference on Learning Representations},
  volume={2024},
  pages={24814--24836},
  year={2024}
}

@article{hu2023radar,
  title={Radar: Robust ai-text detection via adversarial learning},
  author={Hu, Xiaomeng and Chen, Pin-Yu and Ho, Tsung-Yi},
  journal={Advances in neural information processing systems},
  volume={36},
  pages={15077--15095},
  year={2023}
}

@inproceedings{yang2024dnagpt,
  title={Dna-gpt: Divergent n-gram analysis for training-free detection of gpt-generated text},
  author={Yang, Xianjun and Cheng, Wei and Wu, Yue and Petzold, Linda and Wang, William and Chen, Haifeng},
  booktitle={International Conference on Learning Representations},
  volume={2024},
  pages={48572--48597},
  year={2024}
}

@article{hans2024spotting,
  title={Spotting llms with binoculars: Zero-shot detection of machine-generated text},
  author={Hans, Abhimanyu and Schwarzschild, Avi and Cherepanova, Valeriia and Kazemi, Hamid and Saha, Aniruddha and Goldblum, Micah and Geiping, Jonas and Goldstein, Tom},
  journal={arXiv preprint arXiv:2401.12070},
  year={2024}
}

@inproceedings{kirchenbauer2023watermark,
  title={A watermark for large language models},
  author={Kirchenbauer, John and Geiping, Jonas and Wen, Yuxin and Katz, Jonathan and Miers, Ian and Goldstein, Tom},
  booktitle={International conference on machine learning},
  pages={17061--17084},
  year={2023},
  organization={PMLR}
}

@inproceedings{nicks2024language,
  title={Language model detectors are easily optimized against},
  author={Nicks, Charlotte and Mitchell, Eric and Rafailov, Rafael and Sharma, Archit and Manning, Christopher and Finn, Chelsea and Ermon, Stefano},
  booktitle={International Conference on Learning Representations},
  volume={2024},
  pages={7807--7826},
  year={2024}
}

@article{cheng2026adversarial,
  title={Adversarial Paraphrasing: A Universal Attack for Humanizing AI-Generated Text},
  author={Cheng, Yize and Sadasivan, Vinu Sankar and Saberi, Mehrdad and Saha, Shoumik and Feizi, Soheil},
  journal={Advances in Neural Information Processing Systems},
  volume={38},
  pages={47591--47622},
  year={2026}
}

@article{david2025authormist,
  title={Authormist: Evading ai text detectors with reinforcement learning},
  author={David, Isaac and Gervais, Arthur},
  journal={arXiv preprint arXiv:2503.08716},
  year={2025}
}

@inproceedings{creo2025silverspeak,
  title={SilverSpeak: evading AI-generated text detectors using homoglyphs},
  author={Creo, Aldan and Pudasaini, Shushanta},
  booktitle={Proceedings of the 1stWorkshop on GenAI Content Detection (GenAIDetect)},
  pages={1--46},
  year={2025}
}

@article{ranganath2026stealthrl,
  title={StealthRL: Reinforcement Learning Paraphrase Attacks for Multi-Detector Evasion of AI-Text Detectors},
  author={Ranganath, Suraj and Ramesh, Atharv},
  journal={arXiv preprint arXiv:2602.08934},
  year={2026}
}

@article{lu2023large,
  title={Large language models can be guided to evade ai-generated text detection},
  author={Lu, Ning and Liu, Shengcai and He, Rui and Wang, Qi and Ong, Yew-Soon and Tang, Ke},
  journal={arXiv preprint arXiv:2305.10847},
  year={2023}
}

@book{altman1999constrained,
  title     = {Constrained Markov Decision Processes},
  author    = {Altman, Eitan},
  year      = {1999},
  publisher = {Chapman \& Hall/CRC},
  address   = {Boca Raton, FL},
  series    = {Stochastic Modeling}
}

@inproceedings{achiam2017constrained,
  title={Constrained policy optimization},
  author={Achiam, Joshua and Held, David and Tamar, Aviv and Abbeel, Pieter},
  booktitle={International conference on machine learning},
  pages={22--31},
  year={2017},
  organization={Pmlr}
}

@article{krishna2023paraphrasing,
  title={Paraphrasing evades detectors of ai-generated text, but retrieval is an effective defense},
  author={Krishna, Kalpesh and Song, Yixiao and Karpinska, Marzena and Wieting, John and Iyyer, Mohit},
  journal={Advances in neural information processing systems},
  volume={36},
  pages={27469--27500},
  year={2023}
}

@article{sadasivan2025can,
  title={Can AI-generated text be reliably detected? stress testing AI text detectors under various attacks},
  author={Sadasivan, Vinu Sankar and Kumar, Aounon and Balasubramanian, Sriram and Wang, Wenxiao and Feizi, Soheil},
  journal={Transactions on Machine Learning Research},
  year={2025}
}

@article{tessler2019reward,
  title={Reward constrained policy optimization},
  author={Tessler, Chen and Mankowitz, Daniel J and Mannor, Shie},
  journal={arXiv preprint arXiv:1805.11074},
  year={2018}
}

@article{guo2023hc3,
  title={How close is chatgpt to human experts? comparison corpus, evaluation, and detection},
  author={Guo, Biyang and Zhang, Xin and Wang, Ziyuan and Jiang, Minqi and Nie, Jinran and Ding, Yuxuan and Yue, Jianwei and Wu, Yupeng},
  journal={arXiv preprint arXiv:2301.07597},
  year={2023}
}

@article{yu2025peerreview,
  title={Is your paper being reviewed by an LLM? A new benchmark dataset and approach for detecting AI text in peer review},
  author={Yu, Sungduk and Luo, Man and Madusu, Avinash and Lal, Vasudev and Howard, Phillip},
  journal={arXiv e-prints},
  pages={arXiv--2502},
  year={2025}
}

@article{gambetti2023combat,
  title={Combat ai with ai: Counteract machine-generated fake restaurant reviews on social media},
  author={Gambetti, Alessandro and Han, Qiwei},
  journal={arXiv preprint arXiv:2302.07731},
  year={2023}
}

@article{chen2025coconuts,
  title={CoCoNUTS: Concentrating on Content while Neglecting Uninformative Textual Styles for AI-Generated Peer Review Detection},
  author={Chen, Yihan and Chen, Jiawei and Mo, Guozhao and Chen, Xuanang and He, Ben and Han, Xianpei and Sun, Le},
  journal={arXiv preprint arXiv:2509.04460},
  year={2025}
}

@article{guo2024detective,
  title={Detective: Detecting ai-generated text via multi-level contrastive learning},
  author={Guo, Xun and Zhang, Shan and He, Yongxin and Zhang, Ting and Feng, Wanquan and Huang, Haibin and Ma, Chongyang},
  journal={Advances in Neural Information Processing Systems},
  volume={37},
  pages={88320--88347},
  year={2024}
}

@article{wu2025survey,
  title={A survey on llm-generated text detection: Necessity, methods, and future directions},
  author={Wu, Junchao and Yang, Shu and Zhan, Runzhe and Yuan, Yulin and Chao, Lidia Sam and Wong, Derek Fai},
  journal={Computational Linguistics},
  volume={51},
  number={1},
  pages={275--338},
  year={2025}
}

@inproceedings{yang2024survey,
  title={A survey on detection of llms-generated content},
  author={Yang, Xianjun and Pan, Liangming and Zhao, Xuandong and Chen, Haifeng and Petzold, Linda Ruth and Wang, William Yang and Cheng, Wei},
  booktitle={Findings of the Association for Computational Linguistics: EMNLP 2024},
  pages={9786--9805},
  year={2024}
}

@article{kumarage2024survey,
  title={A survey of ai-generated text forensic systems: Detection, attribution, and characterization},
  author={Kumarage, Tharindu and Agrawal, Garima and Sheth, Paras and Moraffah, Raha and Chadha, Aman and Garland, Joshua and Liu, Huan},
  journal={arXiv preprint arXiv:2403.01152},
  year={2024}
}

@article{kehkashan2025ai,
  title={AI-generated text detection: A comprehensive review of methods, datasets, and applications},
  author={Kehkashan, Tanzila and Riaz, Raja Adil and Al-Shamayleh, Ahmad Sami and Akhunzada, Adnan and Ali, Noman and Hamza, Muhammad and Akbar, Faheem},
  journal={Computer Science Review},
  volume={58},
  pages={100793},
  year={2025},
  publisher={Elsevier}
}

@article{liu2025robustness,
  title={Enhancing the Robustness of AI-Generated Text Detectors: A Survey},
  author={Liu, Xin and Li, Yang and Li, Kan},
  journal={Mathematics},
  volume={13},
  number={13},
  pages={2145},
  year={2025},
  publisher={MDPI}
}

@inproceedings{huang2024robust,
  title={Are ai-generated text detectors robust to adversarial perturbations?},
  author={Huang, Guanhua and Zhang, Yuchen and Li, Zhe and You, Yongjian and Wang, Mingze and Yang, Zhouwang},
  booktitle={Proceedings of the 62nd Annual Meeting of the Association for Computational Linguistics (Volume 1: Long Papers)},
  pages={6005--6024},
  year={2024}
}

@inproceedings{zhou2024humanizing,
  title={Humanizing machine-generated content: evading AI-text detection through adversarial attack},
  author={Zhou, Ying and He, Ben and Sun, Le},
  booktitle={Proceedings of the 2024 Joint International Conference on Computational Linguistics, Language Resources and Evaluation (LREC-COLING 2024)},
  pages={8427--8437},
  year={2024}
}

@article{shao2024deepseekmath,
  title={Deepseekmath: Pushing the limits of mathematical reasoning in open language models},
  author={Shao, Zhihong and Wang, Peiyi and Zhu, Qihao and Xu, Runxin and Song, Junxiao and Bi, Xiao and Zhang, Haowei and Zhang, Mingchuan and Li, YK and Wu, Yang and others},
  journal={arXiv preprint arXiv:2402.03300},
  year={2024}
}

@software{vonwerra2020trl,
  title   = {{TRL: Transformers Reinforcement Learning}},
  author  = {von Werra, Leandro and Belkada, Younes and Tunstall, Lewis and Beeching, Edward and Thrush, Tristan and Lambert, Nathan and Huang, Shengyi and Rasul, Kashif and Gallouédec, Quentin},
  license = {Apache-2.0},
  url     = {https://github.com/huggingface/trl},
  year    = {2020}
}

@article{zheng2023judging,
  title={Judging llm-as-a-judge with mt-bench and chatbot arena},
  author={Zheng, Lianmin and Chiang, Wei-Lin and Sheng, Ying and Zhuang, Siyuan and Wu, Zhanghao and Zhuang, Yonghao and Lin, Zi and Li, Zhuohan and Li, Dacheng and Xing, Eric and others},
  journal={Advances in neural information processing systems},
  volume={36},
  pages={46595--46623},
  year={2023}
}

@inproceedings{liu2023g,
  title={G-eval: NLG evaluation using gpt-4 with better human alignment},
  author={Liu, Yang and Iter, Dan and Xu, Yichong and Wang, Shuohang and Xu, Ruochen and Zhu, Chenguang},
  booktitle={Proceedings of the 2023 conference on empirical methods in natural language processing},
  pages={2511--2522},
  year={2023}
}

@article{hu2022lora,
  title={Lora: Low-rank adaptation of large language models.},
  author={Hu, Edward J and Shen, Yelong and Wallis, Phillip and Allen-Zhu, Zeyuan and Li, Yuanzhi and Wang, Shean and Wang, Liang and Chen, Weizhu and others},
  journal={Iclr},
  volume={1},
  number={2},
  pages={3},
  year={2022}
}

\clearpage
\appendix

\section{Experimental Details}

\subsection{Pseudocode of DEPO}\label{sec:pseudocode}
In \Cref{{alg:depo}}, we present the complete pseudocode of DEPO. 
\begin{algorithm}[!htb]
\caption{Detector Evasion Policy Optimization}
\label{alg:depo}
\begin{algorithmic}[1]

\Require Dataset $\mathcal{D}$; 
MAGE detector $P_{\mathrm{detector}}$; 
BERTScore function; 
group size $G$; 
semantic threshold $\tau_{\mathrm{sem}}$; 
PPO clip coefficient $\epsilon$; 
KL coefficient $\beta$; 
step sizes $\eta_\theta, \eta_\lambda$; 
reference policy $\pi_{\mathrm{ref}}$

\State Initialize policy parameters $\theta$
\State Initialize dual variable $\lambda_0 \geq 0$
\State Set behavior policy parameters $\theta_{\mathrm{old}} \gets \theta$

\For{$t = 0, 1, 2, \ldots$}

    \State Sample input text $x \sim \mathcal{D}$

    \State Sample $G$ rewrites
    $\{y_i\}_{i=1}^{G} \sim \pi_{\theta_{\mathrm{old}}}(\cdot \mid x)$

    \For{$i = 1$ to $G$}

        \State Compute detector and semantic rewards $r_{\mathrm{det}}^{(i)}$, $r_{\mathrm{sem}}^{(i)}$

    \EndFor

    \State Compute statistics
    $\mu_{\mathrm{det}}, \sigma_{\mathrm{det}},
    \mu_{\mathrm{sem}}, \sigma_{\mathrm{sem}}$

    \For{$i = 1$ to $G$}

        \State Compute detector advantage
        $
        A_{\mathrm{det}}^{(i)}$

        \State Compute semantic advantage
        $       A_{\mathrm{sem}}^{(i)}
        $

        \State Compute combined advantage:
        $$
        A_{\mathrm{DEPO}}^{(i)}
        =
        A_{\mathrm{det}}^{(i)}
        +
        \lambda_t A_{\mathrm{sem}}^{(i)}
        $$

    \EndFor

    \State Update policy parameters
    \[
    \theta
    \gets
    \theta
    +
    \eta_\theta
    \nabla_\theta
    \mathcal{L}_{\mathrm{policy}}(\theta)
    \]

    \State Update dual variable
    \[
    \lambda_{t+1}
    =
    \max\!\left(
        0,
        \lambda_t
        -
        \eta_\lambda
        \left(
            \mu_{sem}
            -
            \tau_{\mathrm{sem}}
        \right)
    \right)
    \]

    \State Set $\theta_{\mathrm{old}} \gets \theta$

\EndFor

\State \Return $\theta$

\end{algorithmic}
\end{algorithm}

\subsection{Baselines, Detectors, Evaluation Metrics and Evaluation Dataset}
\label{sec:baselineintro}
\noindent\textbf{Baselines.} Below, we briefly introduce the baselines used in our experiments.
\begin{itemize}[leftmargin=*, itemsep=0pt, topsep=0pt, parsep=0pt, partopsep=0pt]
    \item \textit{Direct Paraphrasing}~\citep{krishna2023paraphrasing}:  prompt-based paraphrasing baseline that rewrites the input without detector-specific optimization.
    \item \textit{AuthorMist}~\citep{david2025authormist}: RL-based paraphraser trained with detector feedback to reduce AI-text detectability.
    \item \textit{StealthRL}~\citep{ranganath2026stealthrl}: RL-based paraphraser optimized for detector evasion.
    \item \textit{SilverSpeak}~\citep{creo2025silverspeak}: homoglyph-based attack that perturbs surface characters to confuse AI-text detectors. We use a homoglyph substitution rate of $0.3$.
    \item \textit{Adversarial Paraphrase}~\citep{cheng2026adversarial}: a training-free adversarial paraphrasing attack that uses detector scores to guide candidate selection. We implement it with Qwen3-4B-Instruct-2507, sampling $4$ candidate paraphrases per input with temperature $0.9$ and top-$p=0.95$.
    \item \textit{Single-objective ablations}: \textsc{BERTScore 0} optimizes detector evasion only, while \textsc{BERTScore 1} optimizes semantic preservation only.
    \item \textit{Linear scalarization}: \textsc{BERTScore Linear} combines detector and semantic rewards with fixed weights, testing whether a static reward mixture can recover the constrained trade-off achieved by DEPO.
\end{itemize}

\noindent\textbf{Detectors.} We then introduce the detectors used in our experiments.
\begin{itemize}[leftmargin=*, itemsep=1pt, topsep=2pt, parsep=0pt, partopsep=0pt]
    \item \textit{MAGE}~\citep{li2024mage}:  supervised detector used as the training reward model.
    \item \textit{RoBERTa}~\citep{solaiman2019release}: OpenAI-RoBERTa-Large detector: supervised classifier fine-tuned to distinguish GPT-2 from human.
    \item \textit{RADAR}~\citep{hu2023radar}:  adversarially trained detector designed to improve robustness against paraphrasing attacks.
    \item \textit{Binoculars}~\citep{hans2024spotting}:  zero-shot likelihood-ratio detector that compares the behavior of two language models.
    \item \textit{Fast-DetectGPT}~\citep{bao2024fastdetectgpt}:  zero-shot detector based on conditional probability curvature.
\end{itemize}

\noindent\textbf{Evaluation metrics.} We next introduce the details of evaluation metrics used in our experiments.
\begin{itemize}[leftmargin=*, itemsep=0pt, topsep=0pt, parsep=0pt, partopsep=0pt]
    \item \textit{Detector reward}: $1-P_{\text{detector}}(y)$. %\zou{can we simply use $P_{detector}$, do we have to introduce a new notation?} \wang{Fixed}
    \item \textit{Semantic reward}: {\makebox[0pt][l] BERTScore-F1 between the original text $x$ and the rewritten text $y$, computed with \texttt{microsoft/deberta-xlarge-mnli} as the scoring backbone.}
    \item \textit{Constraint fulfillment}: whether the empirical average semantic reward satisfies the target constraint, i.e.,
    $\mathbb{E}[\mathrm{BERTScore}(x,y)] \geq \tau_{\mathrm{sem}}$.
    % \item \textbf{ASR@$\tau$ / ASR}: the fraction of attacked AI texts classified as human-written, computed as $\mathbb{I}[s(y)<\tau]$. For supervised probability-based detectors, including MAGE, RoBERTa, and RADAR, we use the standard threshold $\tau=0.5$. For zero-shot detectors, whose raw scores are not calibrated probabilities, we select $\tau$ on a held-out validation set by maximizing Youden's $J$ statistic. \zou{this is not clear, need to discuss}
    % \wang{Add new version below}
    \item \textit{ASR@$\tau$ / ASR}: the fraction of attacked AI texts that are classified as human under the decision threshold $\tau$. Since attacked AI texts are treated as the positive AI class, ASR is the complement of the detector's true positive rate:
    $\mathrm{ASR}@\tau = 1-\mathrm{TPR}@\tau$.
    For supervised neural detectors, including MAGE, RoBERTa, and RADAR, we use the standard probability threshold $\tau=0.5$. For zero-shot detectors, whose raw scores are not calibrated probabilities, we select $\tau$ using 500 AI texts and 500 human texts from the corresponding dataset, choosing the threshold that maximizes Youden's $J$ statistic.
    \item \textit{Attack success rate (ASR)@1\%FPR}: the fraction of attacked AI texts classified as human-written at a stricter operating point. For each detector, the threshold is computed from 10,000 human-written samples so that the empirical false positive rate is 1\%, and is fixed when evaluating attacked AI-written texts.
    % \item \textit{$\Delta$AUC}: the area under the receiver operating characteristic curve between human-written texts and attacked AI-written texts. Lower AUC indicates weaker detector separability after attack.
    \item \textit{$\Delta$AUC}: the absolute deviation of the area under the receiver operating characteristic curve from random guessing: $ \Delta \mathrm{AUC} = \left|\mathrm{AUC} - 0.5\right|$.
    Here, $\mathrm{AUC}$ is the area under the receiver operating characteristic curve.
    A smaller $\Delta$AUC indicates that the detector is closer to random guessing and thus has weaker separability between human-written and attacked AI-written texts. 
    We use $\Delta$AUC rather than AUC because an AUC below $0.5$ still indicates separability with a reversed decision direction, rather than complete detector failure.
    % \item \textit{Output length}: average length of rewritten texts, reported as an auxiliary indicator of deletion, compression, or expansion during paraphrasing.
\end{itemize}

\noindent\textbf{Evaluation dataset.} All reported results in our experiments are obtained on evaluation sets that are either officially provided or separately sampled from the available data splits. Specifically, for the Mage Dataset, we use the official validation split as the evaluation set. For the M4 dataset, we adopt the $\textit{dev\_monolingual}$ split setting in Subtask A for evaluation. Since the RAID dataset does not provide a labeled evaluation split, we construct the evaluation set by randomly sampling instances from the training split using a random seed different from that used during training. For the Peer Review dataset, we use reviews collected from the ICLR conferences between 2020--2022 as the evaluation set. Unless otherwise specified, each evaluation set consists of 1,600 randomly sampled AI-generated samples. For the Peer Review dataset, due to the substantially longer average review length, we instead randomly sample 640 AI-generated samples for evaluation.

All baselines, detectors, and evaluation datasets used in this work are publicly available, and we use them in accordance with their respective licenses, terms of use, and intended research purposes.

\subsection{Peer-review Dataset}
\label{sec:peer_review_dataset}

The peer-review evaluation uses the AI Peer Review Detection Benchmark introduced by \citet{yu2025peerreview}. The full benchmark contains paired human- and AI-written peer reviews for research papers submitted to major AI conferences. In this work, we use the ICLR portion from 2017--2022, which provides human-written OpenReview reviews together with AI-generated reviews produced for the same papers. This construction is useful for detector-evasion evaluation because the human and AI reviews are grounded in the same manuscript context, reducing topic mismatch and making the detection problem closer to realistic peer-review misuse scenarios.

The dataset is also practically important. Peer review is a high-stakes academic workflow in which reviewers are expected to provide independent expert judgment. If LLM-written reviews become difficult to detect, program chairs and authors may have limited ability to identify low-effort or undisclosed AI-assisted reviewing. Evaluating detector-evasive paraphrasing on this corpus, therefore, tests our method in a long-form, domain-specialized setting where preserving substantive content matters: a successful rewrite should not only evade the detector but also retain the technical claims, criticisms, and recommendations present in the original review.

\subsection{Peer-review RoBERTa Detector Validation}
\label{sec:peer_review_detector_validation}

\begin{table*}[!htbp]
\centering
\small
\setlength{\tabcolsep}{3pt}
\renewcommand{\arraystretch}{1.15}
\caption{Low-FPR performance of our Peer-review RoBERTa detector. The detector is trained on ICLR 2017--2019 reviews and evaluated on held-out ICLR 2020--2022 reviews generated by GPT-4o, Gemini, and Claude. Anchor and Binoculars results are taken from \citet{yu2025peerreview}. FPR and TPR are reported as percentages.}
\label{tab:peer_review_detector_sota}

\begingroup
\fontsize{9pt}{10pt}\selectfont
\begin{tabular}{llcccccc}
\toprule
 & & \multicolumn{2}{c}{\textbf{Target FPR: 0.1\%}}
   & \multicolumn{2}{c}{\textbf{Target FPR: 0.5\%}}
   & \multicolumn{2}{c}{\textbf{Target FPR: 1\%}} \\
\cmidrule(lr){3-4} \cmidrule(lr){5-6} \cmidrule(lr){7-8}
\textbf{Evaluation} & \textbf{Method}
& \textbf{FPR} & \textbf{TPR}
& \textbf{FPR} & \textbf{TPR}
& \textbf{FPR} & \textbf{TPR} \\
\midrule

\multirow{3}{*}{\shortstack[c]{GPT-4o\\Reviews}}
& Anchor        & 0.1 & 63.5          & 0.5 & 83.7          & 1.0 & 88.8 \\
& Binoculars    & 0.2 & 17.1          & 0.6 & 33.6          & 1.0 & 45.2 \\
& Peer-review RoBERTa  & 0.1 & \textbf{93.2} & 0.4 & \textbf{99.3} & 0.8 & \textbf{99.9} \\
\midrule

\multirow{3}{*}{\shortstack[c]{Gemini\\Reviews}}
& Anchor        & 0.2 & 59.7          & 0.8 & 80.3          & 1.3 & 86.5 \\
& Binoculars    & 0.2 & 61.5          & 0.6 & 78.0          & 1.0 & 85.5 \\
& Peer-review RoBERTa  & 0.1 & \textbf{95.8} & 0.4 & \textbf{99.8} & 0.8 & \textbf{99.9} \\
\midrule

\multirow{3}{*}{\shortstack[c]{Claude\\Reviews}}
& Anchor        & 0.1 & 59.6          & 0.5 & 75.8          & 1.0 & 81.8 \\
& Binoculars    & 0.2 & 43.5          & 0.6 & 65.8          & 1.0 & 77.0 \\
& Peer-review RoBERTa  & 0.1 & \textbf{98.9} & 0.4 & \textbf{99.8} & 0.8 & \textbf{99.8} \\
\bottomrule
\end{tabular}
\endgroup

\end{table*}

For the peer-review evaluation, we additionally train a RoBERTa-based detector on ICLR 2017--2019 reviews from the AI Peer Review Detection Benchmark~\citep{yu2025peerreview}. The detector is evaluated on held-out ICLR 2020--2022 reviews, which tests temporal generalization within the peer-review domain. As shown in Table~\ref{tab:peer_review_detector_sota}, this detector achieves strong low-FPR performance and outperforms the strongest reported baseline detector, Anchor~\cite{yu2025peerreview}, from the original benchmark. We therefore use it as a strong in-domain detector for evaluating detector-evasive paraphrasing on peer-review text.

\subsection{Choice of $\tau_{\mathrm{sem}}$}\label{sec:thresholdtausem}
 The threshold $\tau_{\mathrm{sem}}$ specifies the semantic-preservation region within which detector evasion is optimized. In practice, it can be selected by calibrating BERTScore on a held-out set of ordinary paraphrases and choosing a lower quantile of their similarity scores, so that standard meaning-preserving rewrites are treated as feasible while rewrites with substantial semantic drift are rejected. Following this principle, we set $\tau_{\mathrm{sem}}=0.85$ in our main experiments. This value places direct paraphrasing in the feasible region but rules out detector-only and low-semantic-weight optimization variants that achieve stronger evasion by sacrificing meaning. We use the same threshold across all main evaluations and explicitly report constraint fulfillment, so that attack effectiveness is interpreted solely in terms of whether the semantic constraint is satisfied. 
 
\section{Additional Experiments}
\label{sec:additional_exp_appendix}
In this section, we include additional experiment results and ablation studies that are not discussed in the main paper.

\subsection{Fine-grained Semantic Sensitivity Analysis}
\label{sec:fine_grained_semantic_sensitivity}

\noindent\textbf{Motivation and Scope.}
Semantic preservation is central to detector-evasive paraphrasing. A useful semantic metric should assign high scores to meaning-preserving rewrites, but should also decrease when a small local edit changes the meaning of the sentence. This is important because detector-evasive rewriting may preserve most surface words while altering critical semantic components, such as negation, role structure, causal relations, temporal conditions, or logical constraints.

We state that his analysis is not intended to provide a comprehensive comparison of which metric gives a more general, complete, or universally faithful semantic representation. T5 sentence-level similarity and BERTScore come from different modeling and evaluation paradigms: Sentence-T5 produces fixed-size sentence embeddings compared by cosine similarity, whereas BERTScore performs contextual token-level matching and aggregates token-wise similarities. We therefore treat the following experiment as a task-specific diagnostic: under controlled local semantic perturbations that are especially relevant to detector-evasive paraphrasing, we test whether BERTScore provides a more responsive constraint signal than T5 sentence-level cosine similarity.

\noindent\textbf{Metrics compared.}
We compare sentence-level T5 similarity and BERTScore. For T5 similarity, we use Sentence-T5 embeddings and compute cosine similarity between the source sentence and the candidate rewrite. This provides a global sentence-level similarity score, but may smooth over local meaning changes when the two sentences share most of their lexical content. In contrast, BERTScore computes contextual token-level matching between a candidate and a reference, and aggregates token-wise similarities into precision, recall, and F1~\citep{zhang2020bertscore}. Prior work shows that Sentence-T5 provides strong general-purpose sentence representations~\citep{ni2022sentencet5}, while sentence-level embedding spaces may still be imperfect for fine-grained semantic distinctions such as negation, antonymy, and structural changes~\citep{mahajan2024alignsim}. Related work has also used BERT-family sentence-similarity metrics to evaluate semantic preservation in adversarial paraphrasing; Adversarial attack~\citep{cheng2026adversarial} uses cosine similarity between SBERT embeddings to measure semantic/text-quality preservation before and after paraphrasing. Our goal is therefore not to claim that BERTScore is universally superior, but to test whether it is more responsive to the types of local semantic changes that matter in our constrained paraphrasing setting.

\noindent\textbf{Experimental setup.}
We construct a controlled set of sentence-level perturbation pairs. Starting from manually designed hard cases and automatically generated templates, we first obtain a meaning-preserving rewrite $y^{+}$ for each source sentence $x$ using the paraphrase model with protected-span constraints. We then construct a corrupted rewrite $y^{-}$ by applying a single rule-based semantic perturbation to $y^{+}$. Invalid cases where the rule cannot be applied, does not change the sentence, or produces an ill-formed output are removed. The resulting perturbations are grouped into four categories that focus on structural and logic-sensitive semantic changes: context shifts, logical constraints, logical relations, and role structures.

Given a semantic metric $m$, we compute the score drop caused by the perturbation for each pair as:
\begin{equation}
    \Delta_m^{(i)} = s_m(x_i, y_i^{+}) - s_m(x_i, y_i^{-}),
\end{equation}
where $s_m(\cdot,\cdot)$ denotes the similarity score produced by metric $m$. To make drops easier to interpret within each metric, we also compute the relative score drop:
\begin{equation}
    \mathrm{RelDrop}_m^{(i)} =
    100 \times \frac{s_m(x_i, y_i^{+}) - s_m(x_i, y_i^{-})}
    {|s_m(x_i, y_i^{+})| + \epsilon},
\end{equation}
where $\epsilon$ is a small constant for numerical stability. We report group-level averages of $\Delta_m^{(i)}$ and $\mathrm{RelDrop}_m^{(i)}$. Because T5 cosine similarity and BERTScore-F1 have different score scales, these drops should be interpreted as diagnostic indicators of within-metric sensitivity rather than calibrated cross-metric scores.

\begin{table*}[!h]
\centering
\small
\setlength{\tabcolsep}{3.5pt}
\renewcommand{\arraystretch}{1.15}
\caption{
Diagnostic sensitivity comparison between T5 sentence-level similarity and BERTScore under controlled structural and logic-sensitive semantic perturbations.
$\Delta$ denotes the average absolute score drop from the meaning-preserving rewrite $y^{+}$ to the corrupted rewrite $y^{-}$.
RelDrop denotes the average relative score drop within each metric.
Because the two metrics have different score scales, the drop values should be interpreted as diagnostic indicators rather than calibrated cross-metric comparisons.
}
\label{tab:sensitivity_drop}
\begin{tabular}{lccccc}
\toprule
\textbf{Perturbation group} 
& \textbf{$n$} 
& \textbf{$\Delta$T5} 
& \textbf{$\Delta$BERTScore}
& \textbf{T5 RelDrop} 
& \textbf{BERTScore RelDrop} \\
\midrule
Context shift       & 26  & 0.0197 & 0.0514 & 1.97\% & 5.14\%  \\
Logic constraint    & 51  & 0.0591 & 0.1170 & 5.91\% & 11.70\% \\
Logical relation    & 51  & 0.0387 & 0.0899 & 3.87\% & 8.96\%  \\
Role structure      & 15  & 0.0050 & 0.0605 & 0.50\% & 6.11\%  \\
\midrule
Overall             & 143 & 0.0390 & 0.0895 & 3.90\% & 8.94\%  \\
\bottomrule
\end{tabular}
\end{table*}

\begin{table*}[!h]
\centering
\small
\setlength{\tabcolsep}{3pt}
\renewcommand{\arraystretch}{1.15}
\setlength{\tabcolsep}{5pt}
\caption{
    Representative minimal-edit examples illustrating how T5 sentence-level similarity and BERTScore respond to local semantic perturbations.
    Each example preserves most lexical content while changing a key semantic relation, such as agent--patient roles or condition-specific effectiveness.
}
\label{tab:qualitative_examples}
\begin{tabularx}{\textwidth}{
>{\centering\arraybackslash}p{0.05\textwidth}
>{\raggedright\arraybackslash}X
>{\raggedright\arraybackslash}X
>{\centering\arraybackslash}p{0.15\textwidth}
>{\centering\arraybackslash}p{0.17\textwidth}
}
\toprule
\textbf{No.} 
& \textbf{Reference sentence} 
& \textbf{Perturbed sentence} 
& \textbf{T5 change} 
& \textbf{BERTScore change} \\
\midrule

1
& Alicia gave Marcus the camera before the concert.
& Marcus gave Alicia the camera before the concert.
& $1.0000 \rightarrow 0.9968$ \newline $(-0.32\%)$
& $1.0000 \rightarrow 0.9173$ \newline $(-8.27\%)$ \\
\midrule

2
& Nora beat Liam by three points in the final round.
& Liam beat Nora by three points in the final round.
& $1.0000 \rightarrow 0.9968$ \newline $(-0.32\%)$
& $1.0000 \rightarrow 0.9199$ \newline $(-8.01\%)$ \\
\midrule

3
& The landlord gave the tenant the contract after the inspection.
& The tenant gave the landlord the contract after the inspection.
& $1.0000 \rightarrow 0.9909$ \newline $(-0.91\%)$
& $1.0000 \rightarrow 0.9189$ \newline $(-8.11\%)$ \\
\midrule

4
& The editor gave the writer the draft after the review.
& The writer gave the editor the draft after the review.
& $1.0000 \rightarrow 0.9974$ \newline $(-0.26\%)$
& $1.0000 \rightarrow 0.9315$ \newline $(-6.85\%)$ \\
\midrule

5
& The vaccine was effective in adults but ineffective in children.
& The vaccine was effective in children but ineffective in adults.
& $1.0000 \rightarrow 0.9944$ \newline $(-0.56\%)$
& $1.0000 \rightarrow 0.9343$ \newline $(-6.57\%)$ \\
\bottomrule

\end{tabularx}
\end{table*}

\noindent\textbf{Results and analysis.}
Table~\ref{tab:sensitivity_drop} shows that BERTScore is more responsive than T5 sentence-level similarity to controlled structural and logic-sensitive perturbations in this diagnostic setting. Across 143 constructed pairs, BERTScore has a larger average relative drop than T5 similarity ($8.94\%$ vs. $3.90\%$).

The difference is especially clear for structural changes. For role-structure perturbations, where the agent and patient of an action are swapped while most lexical content remains unchanged, T5 similarity drops by only $0.50\%$, while BERTScore drops by $6.11\%$. BERTScore also shows larger drops for context shifts, logical constraints, and logical relations.

These results suggest that sentence-level embeddings can remain highly similar when most surface words are preserved, even if a local structural or logical edit changes the meaning. In contrast, BERTScore provides a stronger diagnostic signal for the semantic changes emphasized in this work. This does not imply that BERTScore is a universally better semantic representation than T5-based sentence embeddings. Rather, it supports our use of BERTScore as a practical semantic-preservation constraint in DEPO, where the goal is to penalize fine-grained semantic drift under constrained detector-evasive paraphrasing. Representative minimal-edit examples in Table~\ref{tab:qualitative_examples} show the same trend qualitatively.

\subsection{Ablation Study on Different $\tau_{\mathrm{sem}}$}
\label{sec:ablation_tau_sem}
To study the effect of the semantic constraint, we vary the semantic threshold $\tau_{\mathrm{sem}}$ and train DEPO under three different constraint levels. We follow the training framework described in Section~\ref{sec:experimental_setup} and evaluate the resulting models on a randomly selected subset of 640 samples from the MAGE evaluation dataset. Table~\ref{tab:ablation_tau_sem} reports the detector reward and semantic reward obtained under different values of $\tau_{\mathrm{sem}}$.

\begin{table}[!htbp]
\centering
\small
\setlength{\tabcolsep}{6pt}
\renewcommand{\arraystretch}{1.15}
\caption{Ablation study on different semantic constraint thresholds $\tau_{\mathrm{sem}}$.}
\label{tab:ablation_tau_sem}
\begin{tabular}{c c c}
\toprule
\textbf{Target $\tau_{\mathrm{sem}}$} & \textbf{Detector reward} & \textbf{Semantic reward} \\
\midrule
$0.80$ & 0.867 & 0.811 \\
$0.85$ & 0.707 & 0.858 \\
$0.90$ & 0.019 & 0.909 \\
\bottomrule
\end{tabular}
\end{table}

The results reveal a clear trade-off between detector evasion and semantic preservation. A smaller $\tau_{\mathrm{sem}}$ relaxes the semantic constraint and allows more aggressive rewriting, which can improve detector evasion but may weaken semantic preservation. In contrast, a larger $\tau_{\mathrm{sem}}$ imposes a stricter semantic-preservation requirement, encouraging the model to produce rewrites that remain closer in meaning to the source text.

Importantly, our algorithm DEPO is able to adapt to different semantic thresholds and converge under all tested constraint levels. This suggests that the proposed optimization framework can effectively control the balance between detector evasion and semantic preservation by adjusting $\tau_{\mathrm{sem}}$.

\subsection{Detection Score Distribution Analysis}
\label{sec:score_distribution_analysis}

Beyond aggregate ASR and AUC metrics, we also inspect the detector-score distributions induced by different attack methods. This analysis provides a more detailed view of how paraphrasing shifts detector confidence on each evaluation corpus. A successful attack should move AI-written texts toward the human-written score region, while a semantically controlled attack should avoid achieving this shift through excessive deletion or meaning drift.

\begin{figure*}[htbp]
  \centering
  \begin{subfigure}[htbp]{1\textwidth}
    \centering
    \includegraphics[width=\linewidth]{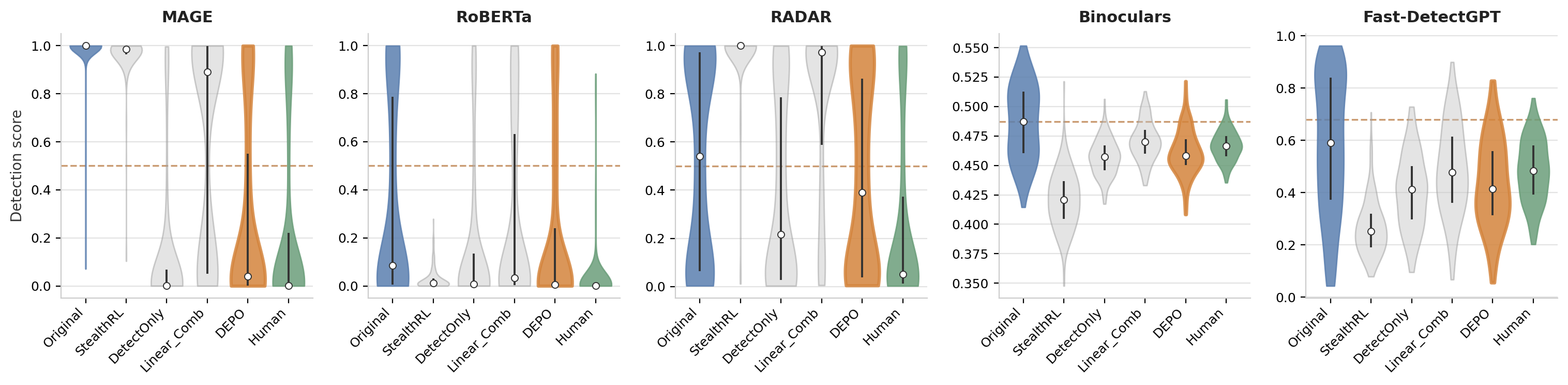}
    \caption{MAGE}
    \label{fig:score-mage}
  \end{subfigure}
  \hfill
  \begin{subfigure}[htbp]{1\textwidth}
    \centering
    \includegraphics[width=\linewidth]{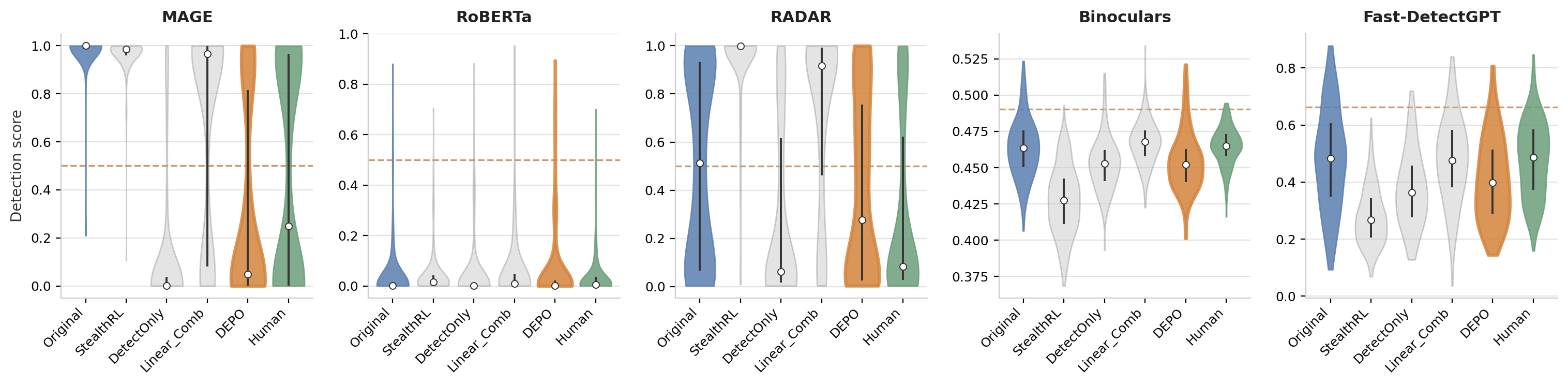}
    \caption{M4}
    \label{fig:score-m4}
  \end{subfigure}
  \vskip0.5em
  \begin{subfigure}[htbp]{1\textwidth}
    \centering
    \includegraphics[width=\linewidth]{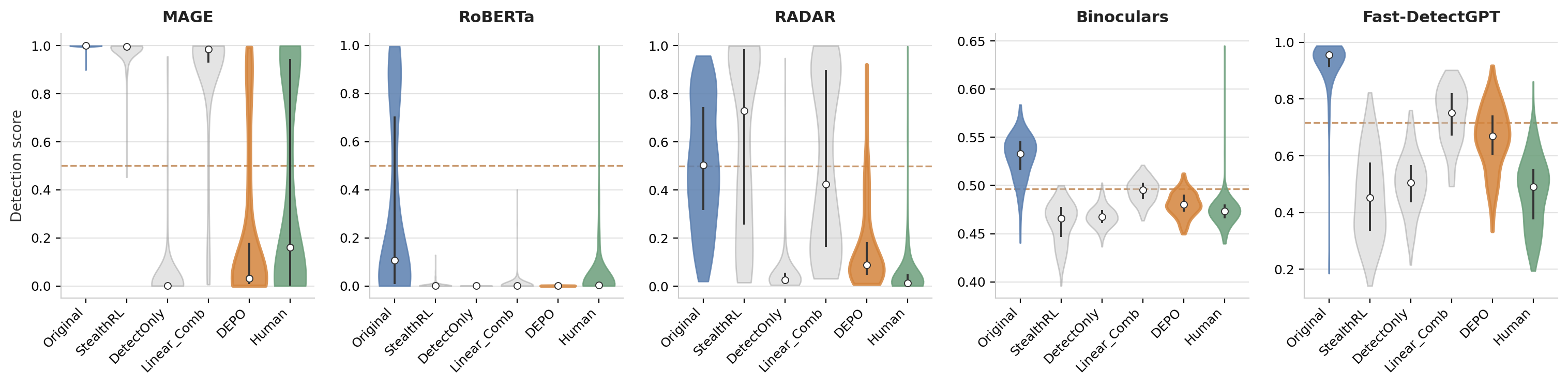}
    \caption{RAID}
    \label{fig:score-raid}
  \end{subfigure}
  \hfill
  \begin{subfigure}[htbp]{1\textwidth}
    \centering
    \includegraphics[width=\linewidth]{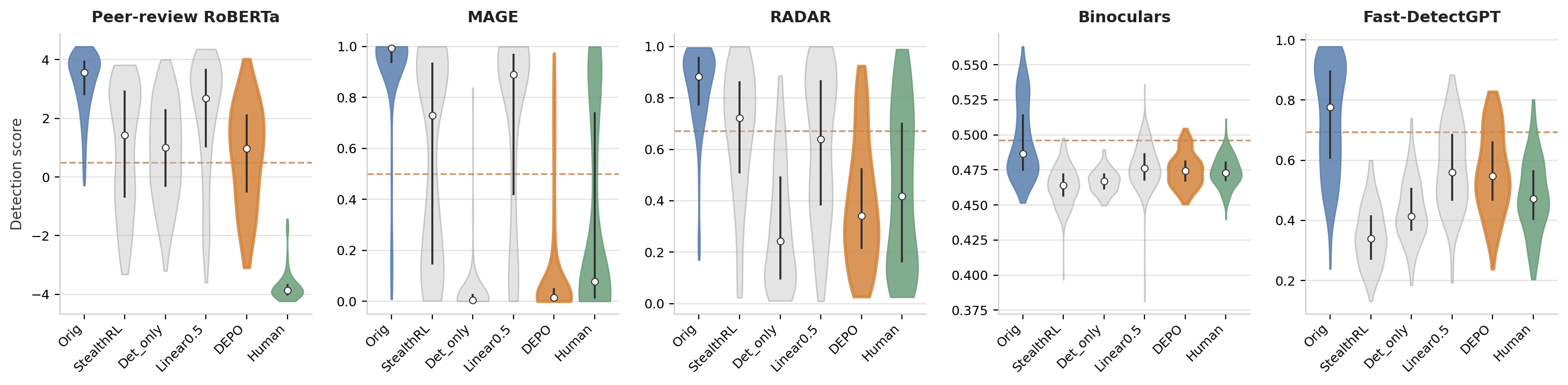}
    \caption{Peer Review}
    \label{fig:score-peer-review}
  \end{subfigure}
    \caption{Detection score distributions across attack methods on four evaluation corpora. This distribution-level analysis complements the aggregate ASR and AUC results in the main text. \texttt{DetectOnly} denotes MAGE BERTScore 0 model, while \texttt{Linear\_Comb} denotes BERTScore Linear 0.5 model.}
  \label{fig:score-all-datasets}
\end{figure*}

Figure~\ref{fig:score-all-datasets} shows the score distributions across MAGE, M4, RAID, and peer-review evaluation sets. The distributions are consistent with the aggregate results in the main text: DEPO shifts detector scores toward lower AI-confidence regions across multiple datasets, while maintaining a more controlled semantic profile than aggressive unconstrained variants. These results complement the main tables by showing that DEPO's gains are not driven only by a small number of threshold-crossing examples, but reflect a broader change in detector-score behavior.

\subsection{Reward Detector Effects on AUC}
\label{sec:reward_detector_auc_analysis}

\begin{figure*}[htbp]
    \centering
    \begin{subfigure}[t]{0.245\textwidth}
        \centering
        \includegraphics[width=\linewidth]{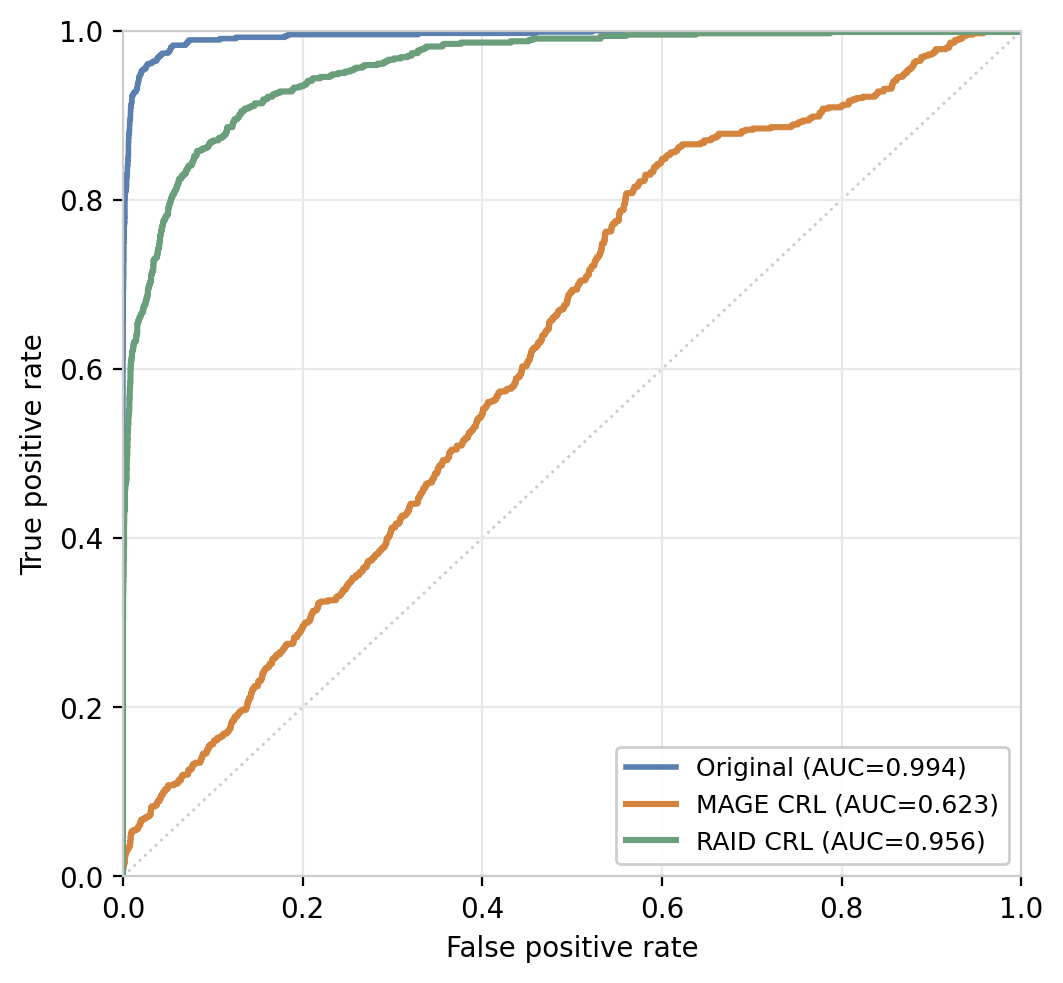}
        \caption{MAGE, MAGE detector}
        \label{fig:mage-mage-detector}
    \end{subfigure}
    \hfill
    \begin{subfigure}[t]{0.245\textwidth}
        \centering
        \includegraphics[width=\linewidth]{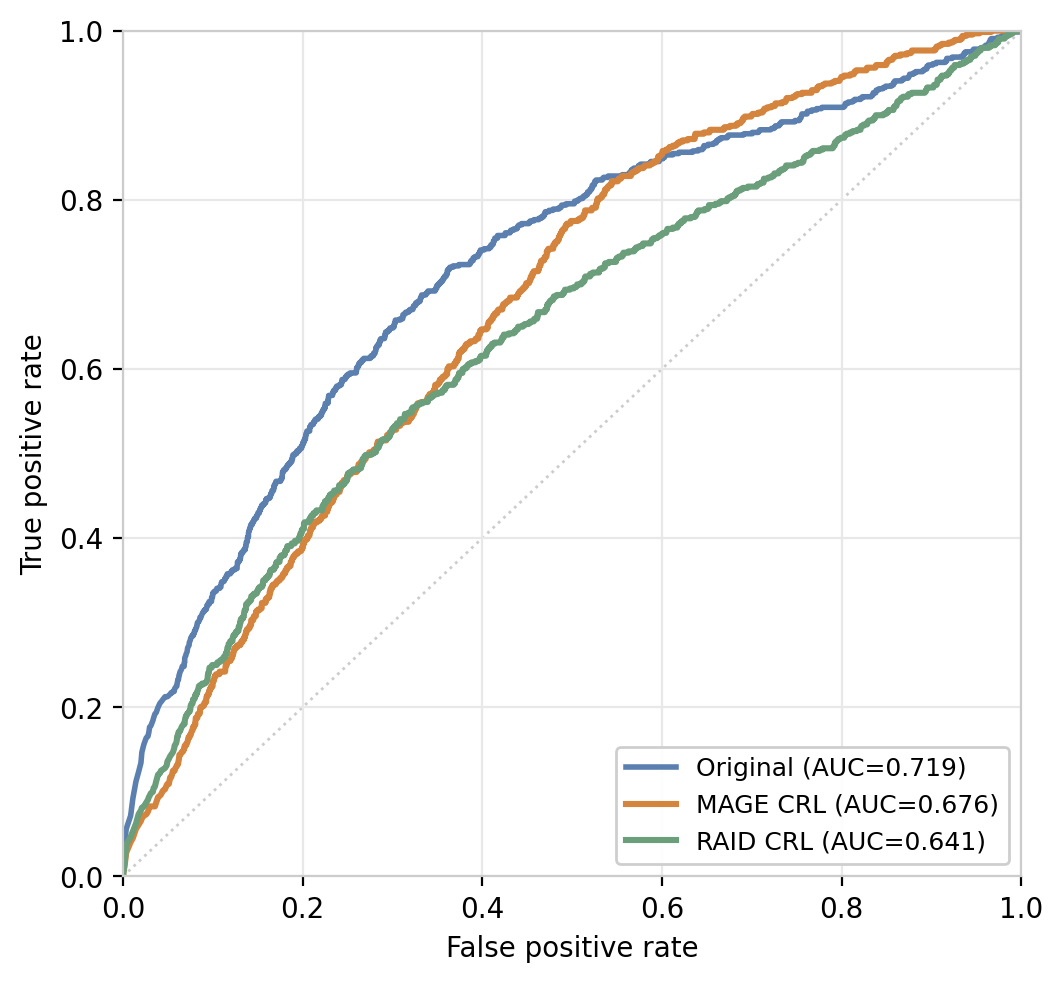}
        \caption{MAGE, RADAR detector}
        \label{fig:mage-radar-detector}
    \end{subfigure}
    \hfill
    \begin{subfigure}[t]{0.245\textwidth}
        \centering
        \includegraphics[width=\linewidth]{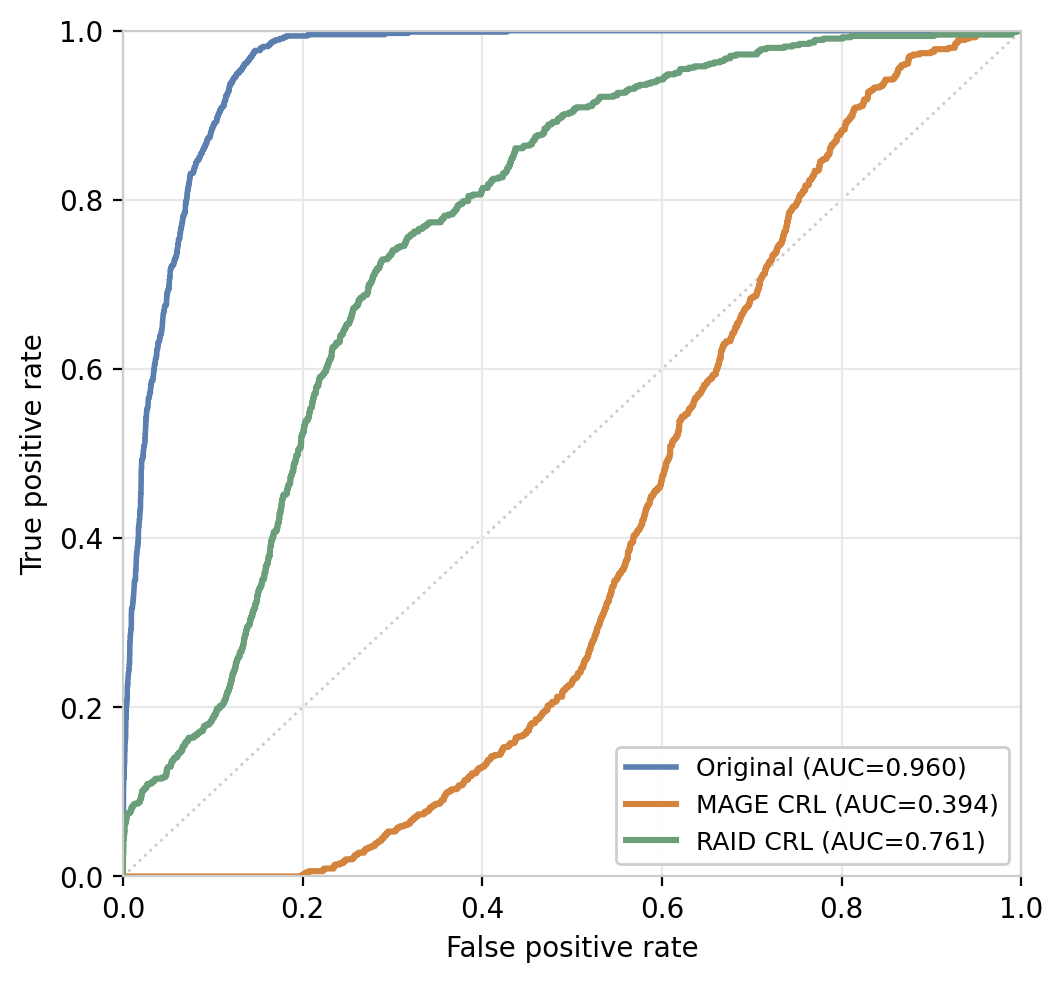}
        \caption{RAID, MAGE detector}
        \label{fig:raid-mage-detector}
    \end{subfigure}
    \hfill
    \begin{subfigure}[t]{0.245\textwidth}
        \centering
        \includegraphics[width=\linewidth]{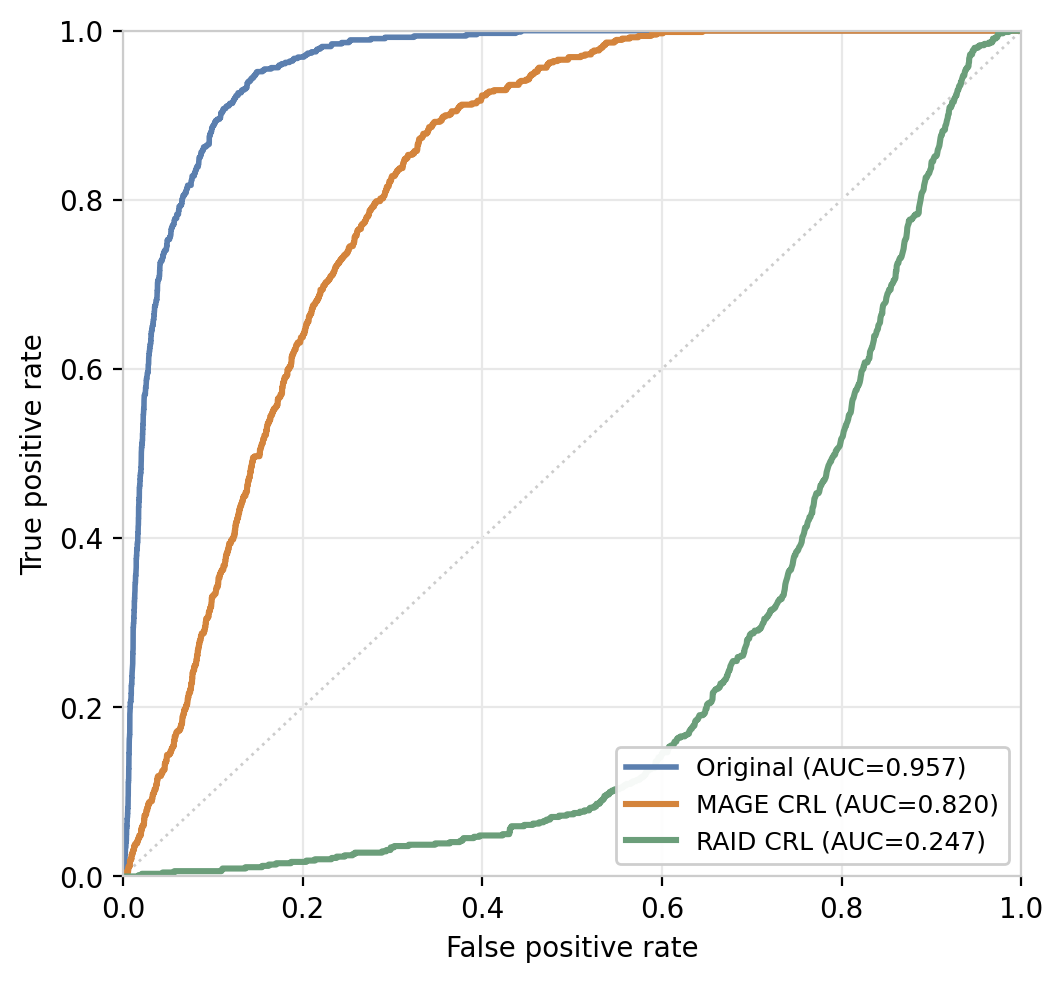}
        \caption{RAID, RADAR detector}
        \label{fig:raid-radar-detector}
    \end{subfigure}

    \caption{
    ROC curves for original texts and DEPO-attacked texts under different reward-detector settings. The policy trained with a given detector reward tends to reduce AUC more strongly on that detector, indicating that detector feedback has a stronger influence on the attack direction than dataset choice alone. In the figure, MAGE CRL denotes the MAGE-trained DEPO policy, while RAID CRL denotes the RADAR-trained DEPO policy.}
    \label{fig:reward-detector-roc-comparison}
\end{figure*}

We further analyze the relative influence of the reward detector and the training dataset on DEPO's generalization behavior. For both the MAGE and RAID datasets, we randomly sample 640 AI-generated texts as rewriting targets and use the same number of human-generated texts to compute ROC curves and AUC. Figure~\ref{fig:reward-detector-roc-comparison} compares ROC curves for policies trained with different detector feedback and evaluated on MAGE and RAID. The main trend is that a policy trained with feedback from a given detector tends to produce stronger confusion for that detector. MAGE-trained DEPO more strongly reduces AUC for the MAGE detector on these two test sets, while RADAR-trained DEPO more strongly reduces AUC for the RADAR detector. This pattern is observed across datasets, suggesting that the reward detector has a larger effect on the learned attack direction than the training dataset.

\subsection{Prompt Robustness Evaluation}
\label{sec:prompt_robustness}
\begin{table*}[!h]
\centering
\small
\setlength{\tabcolsep}{2pt}
\renewcommand{\arraystretch}{1.15}
\caption{Prompt robustness of DEPO on a 640-sample subset of the MAGE evaluation set under different paraphrasing instructions.}
\label{tab:prompt_robustness}
\resizebox{\textwidth}{!}{
\begin{tabular}{lc cc cc cc cc cc}
\toprule
\multirow{2}{*}{\textbf{Prompt Template}}
& \multirow{2}{*}{\makecell{\textbf{BERT}\\\textbf{Score}}}
& \multicolumn{2}{c}{\textbf{MAGE}}
& \multicolumn{2}{c}{\textbf{RoBERTa}}
& \multicolumn{2}{c}{\textbf{RADAR}}
& \multicolumn{2}{c}{\textbf{Binoculars}}
& \multicolumn{2}{c}{\textbf{Fast-DetectGPT}} \\
\cmidrule(lr){3-4}
\cmidrule(lr){5-6}
\cmidrule(lr){7-8}
\cmidrule(lr){9-10}
\cmidrule(lr){11-12}
&
& \textbf{ASR$\uparrow$} & \textbf{$\Delta$AUC$\downarrow$}
& \textbf{ASR$\uparrow$} & \textbf{$\Delta$AUC$\downarrow$}
& \textbf{ASR$\uparrow$} & \textbf{$\Delta$AUC$\downarrow$}
& \textbf{ASR$\uparrow$} & \textbf{$\Delta$AUC$\downarrow$}
& \textbf{ASR$\uparrow$} & \textbf{$\Delta$AUC$\downarrow$} \\
\midrule

Default
& $0.857$
& $0.74$ & $0.14$
& $0.84$ & $0.13$
& $0.57$ & $0.18$
& $0.90$ & $0.11$
& $0.90$ & $0.08$ \\

No Preserving Meaning
& $0.841$
& $0.76$ & $0.12$
& $0.85$ & $0.12$
& $0.62$ & $0.16$
& $0.93$ & $0.14$
& $0.91$ & $0.08$ \\

AuthorMist
& $0.850$
& $0.76$ & $0.13$
& $0.85$ & $0.13$
& $0.59$ & $0.17$
& $0.93$ & $0.12$
& $0.92$ & $0.08$ \\

Reworded
& $0.855$
& $0.73$ & $0.13$
& $0.85$ & $0.14$
& $0.58$ & $0.18$
& $0.91$ & $0.11$
& $0.92$ & $0.08$ \\

Concise
& $0.852$
& $0.75$ & $0.13$
& $0.85$ & $0.13$
& $0.58$ & $0.18$
& $0.92$ & $0.11$
& $0.91$ & $0.08$ \\

\bottomrule
\end{tabular}
}
\end{table*}

We further evaluate whether DEPO relies on a specific paraphrasing instruction. Unlike the main evaluation, which uses 1,600 AI-written texts, this prompt-robustness study is conducted on a 640-sample subset of AI-written texts from the MAGE evaluation set. All prompt templates are evaluated on the same subset using the same detector thresholds as in the main evaluation. Therefore, the absolute numbers in this section are not intended to exactly match the main MAGE results; instead, this analysis focuses on the relative stability of DEPO across different prompt wordings.

Table~\ref{tab:prompt_robustness} shows that DEPO is stable under prompt rewording on the 640-sample evaluation subset. Across five templates, MAGE ASR remains between $0.73$ and $0.76$, while BERTScore stays between $0.841$ and $0.857$. Attacks generated with different prompts are also consistently effective and stable against the other evaluated detectors. Removing the explicit meaning-preservation instruction slightly lowers semantic similarity, but the attack metrics remain close to those of the default prompt. These results suggest that DEPO's behavior is not tied to a single paraphrasing instruction.

\begin{itemize}[leftmargin=*, itemsep=1pt, topsep=2pt, parsep=0pt, partopsep=0pt]
    \item \textbf{Default}: 
    ``Paraphrase the following text while preserving its meaning. Return only the paraphrase.''
    \item \textbf{No preserving meaning}: 
    ``Paraphrase the following text to sound more human-like. Return only the paraphrase.''
    \item \textbf{AuthorMist}: 
    ``Please paraphrase the following text to make it more human-like while preserving the original meaning. Return only the paraphrased text.''
    \item \textbf{Reworded}: 
    ``Rewrite the passage below in your own words, keeping the original meaning intact. Output only the rewritten passage.''
    \item \textbf{Concise}: 
    ``Paraphrase the text below. Output only the paraphrase.''
\end{itemize}

\subsection{Quality Evaluation of Paraphrased Texts}
\label{sec:llm_judge}

Recent work has shown that large language models can serve as scalable automatic judges for open-ended text evaluation, with reasonable agreement with human preferences and flexible rubric-based assessment~\citep{zheng2023judging,liu2023g}. We therefore conduct an LLM-based quality evaluation as a complementary analysis of paraphrase quality. Following the auto-rated quality evaluation protocol of \citet{cheng2026adversarial}, we ask an LLM judge to assign a single 1--5 equivalence rating to each source--paraphrase pair. The rating measures whether the paraphrase preserves the meaning of the original text: a score of 5 indicates that the texts are approximately equivalent, while a score of 1 indicates that they are not topically related.

We use \texttt{gpt-oss-120b} as the judge model. For a fair comparison, all methods are evaluated on the shared subset of 300 AI-written samples selected randomly on MAGE evaluation dataset. Each example is scored independently, and the judge is shown only the source text and the paraphrase without method labels. In addition to the LLM rating, we report BERTScore against the source text and perplexity (PPL) as complementary automatic indicators of semantic overlap and language-model likelihood, respectively. PPL is computed using \texttt{Qwen/Qwen3-4B-Instruct-2507}.

\begin{table}[!h]
\centering
\small
\setlength{\tabcolsep}{2pt}
\renewcommand{\arraystretch}{1.12}
\resizebox{\columnwidth}{!}{
\begin{tabular}{lccc}
\toprule
\textbf{Method} & \textbf{Rating} & \textbf{BERTScore} & \textbf{PPL} \\
\midrule
Original
    & -- & -- & $20.34$ \\
\midrule
Direct Paraphrasing
    & $4.96$ & $0.85$ & $16.25$ \\
AuthorMist
    & $4.64$ & $0.81$ & $26.31$ \\
StealthRL
    & $2.94$ & $0.61$ & $61.60$ \\
SilverSpeak
    & $4.62$ & $0.34$ & $35.12$ \\
Adv. Paraphrase
    & $4.94$ & $0.85$ & $16.44$ \\
MAGE BERTScore 0
    & $4.89$ & $0.80$ & $20.86$ \\
MAGE BERTScore 1
    & $4.99$ & $0.99$ & $19.54$ \\
MAGE BERTScore Linear 0.1
    & $4.86$ & $0.80$ & $23.72$ \\
MAGE BERTScore Linear 0.3
    & $4.93$ & $0.83$ & $20.69$ \\
MAGE BERTScore Linear 0.5
    & $4.90$ & $0.84$ & $17.61$ \\
\midrule
\textbf{DEPO}
    & ${4.94}$ & ${0.86}$ & ${20.60}$ \\
\bottomrule
\end{tabular}
}
\caption{
Quality evaluation of paraphrased texts. Ratings are LLM-based equivalence scores on a 1--5 scale.
}
\label{tab:llm_judge}
\end{table}

\noindent\textbf{Results.}
Table~\ref{tab:llm_judge} summarizes the LLM-judge evaluation results for DEPO and representative attack baselines.
DEPO achieves a high LLM-based rating under the \texttt{gpt-oss-120b} judge. This indicates that, even when evaluated by a larger and stronger external quality rater, DEPO preserves the semantic equivalence and overall quality of the rewritten outputs. The result suggests that the proposed optimization does not obtain detector evasion by sacrificing paraphrase fidelity.

DEPO also yields a higher PPL than Direct Paraphrasing, while remaining close to the original text and other MAGE-based variants. Since Direct Paraphrasing produces very low-PPL outputs, its generations may be overly smoothed by the paraphrasing model. In contrast, DEPO moves the output distribution away from this highly regularized paraphrase style, suggesting more diverse and human-like linguistic variation while maintaining high judged quality.

\begin{figure*}[t]
    \centering
    \includegraphics[
        width=1\linewidth,
        keepaspectratio
    ]{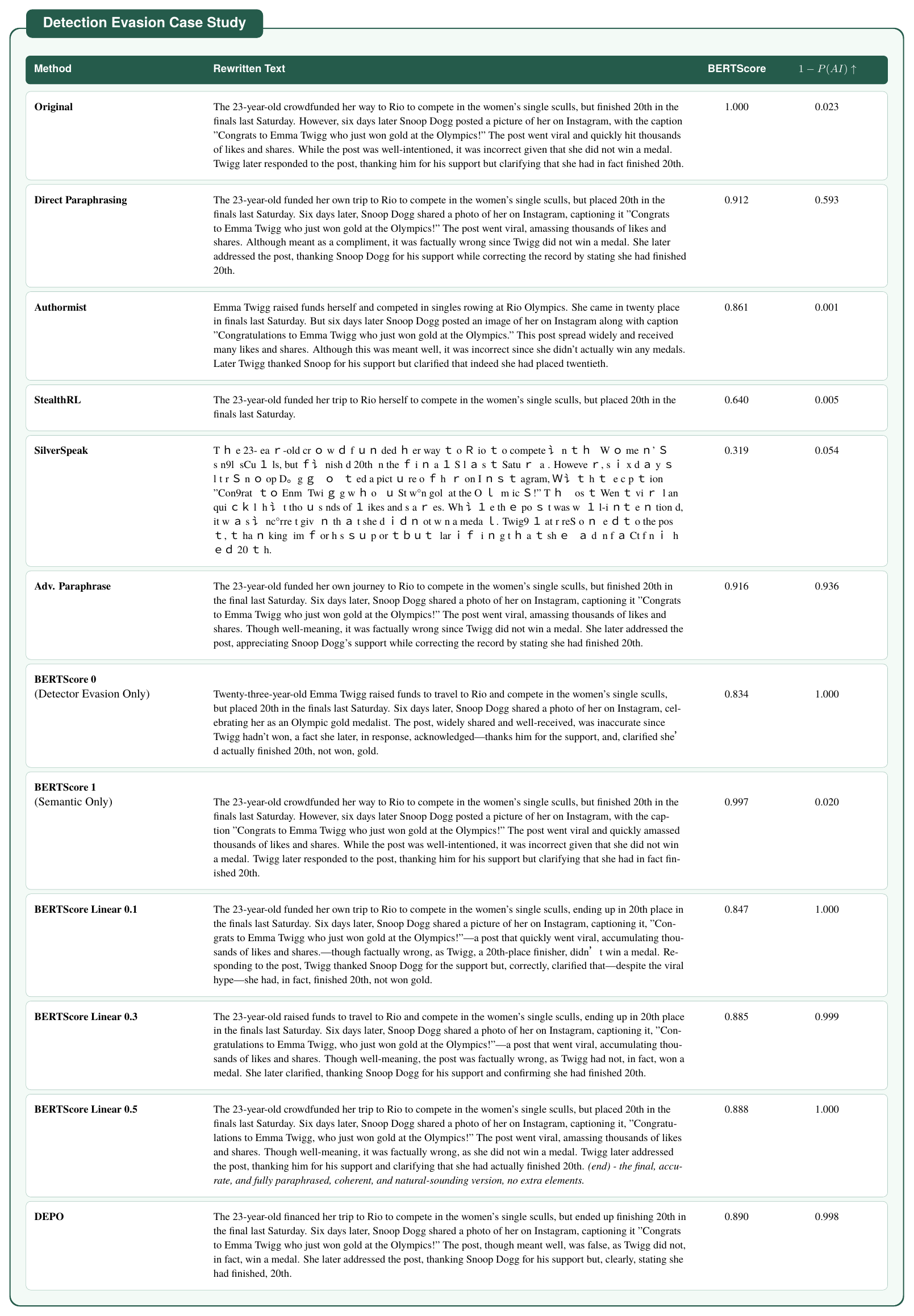}
    \caption{
    \textbf{Qualitative comparison of paraphrasing attack results.}
    This figure presents representative paraphrasing attack outputs for a given AI-generated sentence(Original) using different methods described in Appendix~\ref{sec:baselineintro}.
    }
    \label{fig:detection-evasion-case-study}
\end{figure*}

Finally, the LLM ratings and BERTScore exhibit a broadly consistent trend: methods with higher BERTScore generally receive higher judge ratings, whereas lower BERTScore is often associated with lower equivalence scores. This agreement provides additional evidence that BERTScore is a reasonable proxy for semantic similarity in our setting, and supports its use as the semantic preservation signal in DEPO. A qualitative Example can be found in Fig~\ref{fig:detection-evasion-case-study}.

\noindent\textbf{LLM Judge Prompt Template.}
We use the following prompt template to configure \texttt{gpt-oss-120b} as an LLM-based paraphrase quality judge.

\begin{tcolorbox}[colback=gray!5,colframe=gray!40,title=LLM Judge Prompt]
\small
\textbf{System prompt.}

You are an expert linguist and paraphrase evaluator. Your task is to assess the quality of a paraphrased text compared to the original source text. Use the following scoring criteria:

5 - Approximately equivalent: Meaning is preserved; differences are only in wording or structure.

4 - Nearly equivalent: Meaning is mostly preserved; minor factual details differ.

3 - Somewhat equivalent: Some meaning is preserved; important details or meanings differ.

2 - Topically related: The texts are on the same topic but most meaning is lost.

1 - Not topically related: The texts are not related in topic or meaning.

Provide your final output as a JSON object in this format:

\begin{flushleft}
\ttfamily
\{ "score": \textless{}score from 1 to 5\textgreater{}, "justification": "\textless{}brief explanation\textgreater{}" \}
\end{flushleft}

\vspace{1em}
\textbf{User prompt.}

Evaluate the following paraphrase using the criteria above:

Original Text: \textless{}original\_text\textgreater{}

Paraphrased Text: \textless{}paraphrased\_text\textgreater{}

What score (1 to 5) would you assign to this paraphrase, and why?
\end{tcolorbox}

\subsection{Interactive Demo and License}
We release an interactive demo of our DEPO model on Hugging Face.
Users can input a piece of AI-generated text to obtain its paraphrased version. The demo also reports the MAGE detector probabilities before and after paraphrasing, The demo model is released under the MIT License.

\section{Additional Tables Mentioned in the Main Paper}
\label{sec:table_appendix}
In this section, we include the tables referred to in the main paper.

% ===== Raid =====
\begin{table*}[!htbp]
\centering
\small
\setlength{\tabcolsep}{2pt}
\renewcommand{\arraystretch}{1.15}
\caption{Attack effectiveness on the RAID evaluation set under different AI-text detectors.}
\label{tab:attack_raid}
% \resizebox{\textwidth}{!}{
{
\begin{tabular}{lc cc cc cc cc cc}
\toprule
\multirow{2}{*}{\textbf{Method}}
& \multirow{2}{*}{\makecell{\textbf{BERT}\\\textbf{Score}}}
& \multicolumn{2}{c}{\textbf{MAGE}}
& \multicolumn{2}{c}{\textbf{RoBERTa}}
& \multicolumn{2}{c}{\textbf{RADAR}}
& \multicolumn{2}{c}{\textbf{Binoculars}}
& \multicolumn{2}{c}{\textbf{Fast-DetectGPT}} \\
\cmidrule(lr){3-4}
\cmidrule(lr){5-6}
\cmidrule(lr){7-8}
\cmidrule(lr){9-10}
\cmidrule(lr){11-12}
&
& \textbf{ASR$\uparrow$} & \textbf{$\Delta$AUC$\downarrow$}
& \textbf{ASR$\uparrow$} & \textbf{$\Delta$AUC$\downarrow$}
& \textbf{ASR$\uparrow$} & \textbf{$\Delta$AUC$\downarrow$}
& \textbf{ASR$\uparrow$} & \textbf{$\Delta$AUC$\downarrow$}
& \textbf{ASR$\uparrow$} & \textbf{$\Delta$AUC$\downarrow$} \\
\midrule
\rowcolor{rowgray}
Original
& --
& $0.00$ & $0.46$
& $0.30$ & $0.27$
& $0.10$ & $0.46$
& $0.06$ & $0.48$
& $0.03$ & $0.49$ \\

\midrule
Direct Paraphrasing
& $0.883$
& $0.00$ & $0.38$
& $0.84$ & $0.07$
& $0.35$ & $0.39$
& $0.36$ & $0.42$
& $0.27$ & $0.48$ \\

AuthorMist
& $0.831$
& $0.07$ & $0.31$
& $0.98$ & $0.27$
& $0.78$ & $0.20$
& $0.90$ & $0.07$
& $0.96$ & $0.19$ \\

StealthRL
& $0.730$
& $0.03$ & $0.33$
& $0.94$ & $0.21$
& $0.22$ & $0.43$
& $0.96$ & $0.19$
& $0.97$ & $0.05$ \\

SilverSpeak
& $0.373$
& $0.00$ & $0.33$
& $0.00$ & $0.49$
& $0.00$ & $0.49$
& $0.05$ & $0.49$
& $1.00$ & $0.24$ \\

Adv. Paraphrase
& $0.879$
& $0.00$ & $0.48$
& $0.97$ & $0.25$
& $0.99$ & $0.26$
& $0.58$ & $0.40$
& $0.21$ & $0.48$ \\

\midrule
MAGE BERTScore 0
& $0.821$
& $0.96$ & $0.27$
& $0.98$ & $0.30$
& $0.89$ & $0.14$
& $0.99$ & $0.16$
& $0.99$ & $0.06$ \\

MAGE BERTScore Linear 0.1
& $0.807$
& $0.96$ & $0.26$
& $1.00$ & $0.31$
& $0.63$ & $0.29$
& $0.94$ & $0.13$
& $0.96$ & $0.22$ \\

MAGE BERTScore Linear 0.3
& $0.862$
& $0.86$ & $0.14$
& $0.98$ & $0.30$
& $0.74$ & $0.26$
& $0.91$ & $0.12$
& $0.90$ & $0.29$ \\

\dashedmidrule

MAGE BERTScore 1
& $0.986$
& $0.00$ & $0.45$
& $0.33$ & $0.26$
& $0.12$ & $0.45$
& $0.05$ & $0.48$
& $0.03$ & $0.49$ \\

MAGE BERTScore Linear 0.5
& $0.881$
& $0.12$ & $0.26$
& $0.89$ & $\mathbf{0.10}$
& $0.20$ & $0.44$
& $0.48$ & $0.39$
& $0.50$ & $0.44$ \\

\rowcolor{rowgray}
\textbf{DEPO}
& $0.878$
& $\mathbf{0.86}$ & $\mathbf{0.11}$
& $\mathbf{0.99}$ & ${0.33}$
& $\mathbf{0.65}$ & $\mathbf{0.32}$
& $\mathbf{0.85}$ & $\mathbf{0.15}$
& $\mathbf{0.78}$ & $\mathbf{0.37}$ \\

\bottomrule
\end{tabular}
}
\end{table*}

% ===== PeerReview =====
\begin{table*}[!htbp]
\centering
\small
\setlength{\tabcolsep}{2pt}
\renewcommand{\arraystretch}{1.15}
\caption{Attack effectiveness on the peer-review evaluation set under different AI-text detectors.}
\label{tab:attack_reviews}
% \resizebox{\textwidth}{!}{
{
\begin{tabular}{lc cc cc cc cc cc}
\toprule
\multirow{2}{*}{\textbf{Method}}
& \multirow{2}{*}{\makecell{\textbf{BERT}\\\textbf{Score}}}
& \multicolumn{2}{c}{\makecell{\textbf{PeerReview}\\\textbf{RoBERTa}}}
& \multicolumn{2}{c}{\textbf{MAGE}}
& \multicolumn{2}{c}{\textbf{RADAR}}
& \multicolumn{2}{c}{\textbf{Binoculars}}
& \multicolumn{2}{c}{\textbf{Fast-DetectGPT}} \\
\cmidrule(lr){3-4}
\cmidrule(lr){5-6}
\cmidrule(lr){7-8}
\cmidrule(lr){9-10}
\cmidrule(lr){11-12}
&
& \textbf{ASR$\uparrow$} & \textbf{$\Delta$AUC$\downarrow$}
& \textbf{ASR$\uparrow$} & \textbf{$\Delta$AUC$\downarrow$}
& \textbf{ASR$\uparrow$} & \textbf{$\Delta$AUC$\downarrow$}
& \textbf{ASR$\uparrow$} & \textbf{$\Delta$AUC$\downarrow$}
& \textbf{ASR$\uparrow$} & \textbf{$\Delta$AUC$\downarrow$} \\
\midrule
\rowcolor{rowgray}
Original
& --
& $0.03$ & $0.50$
& $0.09$ & $0.40$
& $0.10$ & $0.35$
& $0.61$ & $0.20$
& $0.39$ & $0.35$ \\

\midrule
Direct Paraphrasing
& $0.818$
& $0.04$ & $0.50$
& $0.07$ & $0.39$
& $0.53$ & $0.13$
& $0.84$ & $0.11$
& $0.64$ & $0.28$ \\

AuthorMist
& $0.721$
& $0.53$ & $0.49$
& $0.32$ & $0.24$
& $0.96$ & $0.27$
& $1.00$ & $0.37$
& $1.00$ & $0.32$ \\

StealthRL
& $0.698$
& $0.40$ & $0.49$
& $0.48$ & $0.17$
& $0.43$ & $0.19$
& $0.99$ & $0.28$
& $1.00$ & $0.33$ \\

SilverSpeak
& $0.392$
& $0.43$ & $0.50$
& $0.00$ & $0.47$
& $0.39$ & $0.22$
& $0.06$ & $0.49$
& $1.00$ & $0.33$ \\

Adv. Paraphrase
& $0.819$
& $0.04$ & $0.50$
& $0.13$ & $0.32$
& $0.58$ & $0.10$
& $0.89$ & $0.09$
& $0.68$ & $0.27$ \\

\midrule
MAGE BERTScore 0
& $0.777$
& $0.44$ & $0.50$
& $0.97$ & $0.29$
& $0.91$ & $0.16$
& $1.00$ & $0.27$
& $0.98$ & $0.17$ \\

MAGE BERTScore Linear 0.1
& $0.737$
& $0.63$ & $0.49$
& $0.98$ & $0.31$
& $0.76$ & $0.06$
& $0.99$ & $0.10$
& $0.98$ & $0.09$ \\

MAGE BERTScore Linear 0.3
& $0.798$
& $0.39$ & $0.49$
& $0.96$ & $0.25$
& $0.90$ & $0.12$
& $0.99$ & $0.13$
& $0.96$ & $0.04$ \\

\dashedmidrule

MAGE BERTScore 1
& $0.913$
& $0.02$ & $0.50$
& $0.07$ & $0.41$
& $0.13$ & $0.34$
& $0.62$ & $0.21$
& $0.39$ & $0.37$ \\

MAGE BERTScore Linear 0.5
& $0.781$
& $0.21$ & $\mathbf{0.49}$
& $0.27$ & $0.24$
& $0.52$ & $0.14$
& $0.88$ & $\mathbf{0.01}$
& $0.77$ & $0.15$ \\

\rowcolor{rowgray}
\textbf{DEPO}
& $0.825$
& $\mathbf{0.41}$ & $\mathbf{0.49}$
& $\mathbf{0.94}$ & $\mathbf{0.23}$
& $\mathbf{0.84}$ & $\mathbf{0.07}$
& $\mathbf{0.95}$ & ${0.06}$
& $\mathbf{0.78}$ & $\mathbf{0.12}$ \\

\bottomrule
\end{tabular}
}
\end{table*}

% ===== RAID =====
\begin{table*}[htbp]
\centering
\small
\setlength{\tabcolsep}{2pt}
\renewcommand{\arraystretch}{1.15}
\caption{Cross-detector evaluation of the RAID-trained, RADAR-targeted DEPO policy on the RAID evaluation set.}
\label{tab:attack_raid_radar_trained}
% \resizebox{\textwidth}{!}{
{
\begin{tabular}{lc cc cc cc cc cc}
\toprule
\multirow{2}{*}{\textbf{Method}}
& \multirow{2}{*}{\makecell{\textbf{BERT}\\\textbf{Score}}}
& \multicolumn{2}{c}{\textbf{MAGE}}
& \multicolumn{2}{c}{\textbf{RoBERTa}}
& \multicolumn{2}{c}{\textbf{RADAR}}
& \multicolumn{2}{c}{\textbf{Binoculars}}
& \multicolumn{2}{c}{\textbf{Fast-DetectGPT}} \\
\cmidrule(lr){3-4}
\cmidrule(lr){5-6}
\cmidrule(lr){7-8}
\cmidrule(lr){9-10}
\cmidrule(lr){11-12}
& 
& \textbf{ASR$\uparrow$} & \textbf{$\Delta$AUC$\downarrow$}
& \textbf{ASR$\uparrow$} & \textbf{$\Delta$AUC$\downarrow$}
& \textbf{ASR$\uparrow$} & \textbf{$\Delta$AUC$\downarrow$}
& \textbf{ASR$\uparrow$} & \textbf{$\Delta$AUC$\downarrow$}
& \textbf{ASR$\uparrow$} & \textbf{$\Delta$AUC$\downarrow$} \\
\midrule
\rowcolor{rowgray}
Original
& --
& $0.00$ & $0.47$ & $0.54$ & $0.35$ & $0.44$ & $0.42$ & $0.23$ & $0.31$ & $0.20$ & $0.32$ \\

\midrule
Direct Paraphrasing
& $0.877$
& $0.01$ & $0.38$ & $0.96$ & $0.07$ & $0.72$ & $0.37$ & $0.50$ & $0.29$ & $0.34$ & $0.32$ \\

AuthorMist
& $0.825$
& $0.09$ & $0.32$ & $0.94$ & $0.07$ & $0.84$ & $0.24$ & $0.91$ & $0.01$ & $0.89$ & $0.09$ \\

StealthRL
& $0.733$
& $0.03$ & $0.34$ & $1.00$ & $0.03$ & $0.26$ & $0.46$ & $0.99$ & $0.31$ & $0.98$ & $0.20$ \\

SilverSpeak
& $0.374$
& $0.00$ & $0.33$ & $0.21$ & $0.48$ & $0.14$ & $0.49$ & $0.25$ & $0.48$ & $1.00$ & $0.23$ \\

Adv. Paraphrase
& $0.876$
& $0.02$ & $0.37$ & $0.98$ & $0.03$ & $0.77$ & $0.35$ & $0.57$ & $0.27$ & $0.41$ & $0.31$ \\

\midrule
RADAR BERTScore 0
& $0.775$
& $0.33$ & $0.19$ & $0.99$ & $0.21$ & $0.98$ & $0.29$ & $0.99$ & $0.28$ & $1.00$ & $0.16$ \\

RADAR BERTScore Linear 0.1
& $0.783$
& $0.34$ & $0.17$ & $0.99$ & $0.21$ & $0.97$ & $0.27$ & $0.99$ & $0.23$ & $0.99$ & $0.14$ \\

\dashedmidrule

RADAR BERTScore 1
& $0.991$
& $0.00$ & $0.47$ & $0.56$ & $0.35$ & $0.46$ & $0.41$ & $0.23$ & $0.32$ & $0.19$ & $0.32$ \\

\rowcolor{rowgray}
\textbf{DEPO}
& $0.850$
& $\mathbf{0.19}$ & $\mathbf{0.27}$ & $\mathbf{0.98}$ & $\mathbf{0.13}$ & $\mathbf{0.95}$ & $\mathbf{0.09}$ & $\mathbf{0.87}$ & $\mathbf{0.04}$ & $\mathbf{0.84}$ & $\mathbf{0.14}$ \\

\bottomrule
\end{tabular}
}
\end{table*}

% \subsection{Hyperparameters and Configuration}
% \label{sec:hyperparameters}
\begin{table*}[!htbp]
\centering
\setlength{\tabcolsep}{2pt}
\renewcommand{\arraystretch}{1.15}
\caption{Complete hyperparameters and configuration.}
\label{tab:hyperparameters}

\begingroup
\small
\begin{tabular}{p{0.28\linewidth} p{0.62\linewidth}}
\toprule
\textbf{Parameter} & \textbf{Value} \\
\midrule

\multicolumn{2}{l}{\textit{Model \& LoRA}} \\
Base model & Qwen/Qwen3-4B-Instruct-2507 \\
LoRA rank & 32 \\
LoRA alpha & 64 \\
LoRA dropout & 0.05 \\
\midrule

\multicolumn{2}{l}{\textit{Training}} \\
Algorithm & DEPO with Lagrangian GRPO-style policy optimization \\
Optimizer & AdamW \\
Adam $\beta_1$, $\beta_2$ & 0.9, 0.999 \\
Learning rate & $2 \times 10^{-4}$ \\
Batch size (prompts) & 4 \\
Gradient accumulation & 4 \\
Group size $G$ & 8 \\
Training epochs & 3 \\
Training samples & 1,000 AI-generated samples \\
Temperature & 1.0 \\
KL penalty coefficient $\beta$ & 0.04 \\
Reference policy & Qwen3-4B-Instruct-2507 (frozen) \\
Hardware & NVIDIA A6000 Ada / H100 GPU \\
Detector reward & Detector $1-P(AI)$ \\
Semantic reward & BERTScore-F1 with microsoft/deberta-xlarge-mnli \\
Radom seed & 42 \\
\midrule

\multicolumn{2}{l}{\textit{Lagrangian \& Constraints}} \\
Target semantic threshold $\tau_{\mathrm{sem}}$ & 0.85 \\
Multiplier init $\lambda_0$ & 0.0 \\
Multiplier LR $\eta_\lambda$ & 0.1 \\
Multiplier max $\lambda_{\max}$ & 10.0 \\
% EMA decay $\rho$ & 0.9 \\
% EMA init $J_0$ & 0.7 \\
\midrule

\multicolumn{2}{l}{\textit{Inference}} \\
Temperature & 0.9 \\
Top-$p$ & 0.95 \\
Max tokens & 512 \\
Default prompt template & ``Paraphrase the following text while preserving its meaning. Return only the paraphrase. Text: [TEXT]'' \\
\midrule

\multicolumn{2}{l}{\textit{Detectors}} \\
MAGE detector & Longformer-based classifier from yaful/MAGE \\
RoBERTa OpenAI & openai-community/roberta-large-openai-detector \\
RADAR & TrustSafeAI/RADAR-Vicuna-7B \\
Binoculars (lightweight) & gpt2-medium + gpt2-large \\
Fast-DetectGPT & Scoring LM: EleutherAI/gpt-neo-2.7B \\
Peer-review RoBERTa & roberta-base model fine-tuned on peer-review dataset \\
\midrule

\multicolumn{2}{l}{\textit{Evaluation}} \\
Main evaluation test samples & 1,600 AI-written samples \\
Token window & 100--500 tokens \\
FPR calibration & 1\% on 10,000 human samples using quantile threshold \\
Candidates per sample & 1 \\
Random seed & 2026 \\
\bottomrule
\end{tabular}
\vspace{-8pt}
\endgroup
\end{table*}

% \subsection{Qualitative Example}

\end{document}